\documentclass{article}

\usepackage[final,nonatbib]{neurips_2021}

\usepackage[utf8]{inputenc} %
\usepackage[T1]{fontenc}    %
\usepackage[colorlinks]{hyperref}       %
\hypersetup{linkcolor=red,citecolor=green,urlcolor=magenta,menucolor=red}
\usepackage{url}            %
\usepackage{booktabs}       %
\usepackage{amsfonts}       %
\usepackage{nicefrac}       %
\usepackage{microtype}      %
\usepackage{xcolor}         %

\usepackage{algorithm}
\usepackage{algorithmic}
\usepackage{amsmath}
\usepackage{changepage}
\usepackage{stackengine}
\usepackage{enumitem}
\usepackage{placeins}
\usepackage{comment}
\usepackage{wrapfig}
\usepackage{dblfloatfix}

\input{math_commands.tex}
\newcommand{\cutparagraphup}{\vspace{-7pt}}

\newcommand{\cutsectionup}{\vspace*{-0.1in}}
\newcommand{\cutsectiondown}{\vspace*{-0.1in}}
\newcommand{\cutsubsectionup}{\vspace*{-0.1in}}
\newcommand{\cutsubsectiondown}{\vspace*{-0.05in}}

\def\fourroom{\textbf{FourRoom}}
\def\multiroom{\textbf{MultiRoom}}
\def\minigrid{\textbf{MiniGrid}}
\def\vizdoom{\textbf{ViZDoom}}

\def\slm{\text{SFL}\xspace}
\def\sfs{\text{SFS}^{\bar{\pi}}}
\def\sf{\psi^{\pi}}
\def\barsf{\psi^{\bar{\pi}}}
\def\randpi{\bar{\pi}}
\def\randsf{\psi^{\randpi}}
\def\randsfs{\text{SFS}^{\randpi}}

\def\landprev{l_{\text{prev}}}
\def\landcurr{l_{\text{curr}}}
\def\landfront{l_{\text{front}}}
\def\landtarget{l_{\text{target}}}

\def\localpi{\pi_l}

\def\eqref#1{equation~\ref{#1}}

\usepackage{cleveref}
\crefname{equation}{Eq.}{}

\usepackage{graphicx}
\graphicspath{ {./images/} }

\usepackage{caption}
\usepackage{subcaption}
\newsavebox{\largestimage}
\usepackage{textcomp}

\title{Successor Feature Landmarks for Long-Horizon Goal-Conditioned Reinforcement Learning}

\author{%
  Christopher Hoang $^{1}$
  \hspace{1em}
  Sungryull Sohn $^{1\ 2}$
  \hspace{1em}
  Jongwook Choi $^{1}$
  \hspace{1em}\\ \bf
  Wilka Carvalho $^{1}$
  \hspace{1em}
  Honglak Lee $^{1\ 2}$
  \\[2pt]
  $^1$University of Michigan \hspace{2em}
  $^2$LG AI Research \hspace{2em}
  \\
  \texttt{{\rm\{}%
        choang, srsohn, jwook, wcarvalh, honglak%
    {\rm\}}@umich.edu} \\
}

\begin{document}

\maketitle

\begin{abstract}

Operating in the real-world often requires agents to learn about a complex environment and apply this understanding to achieve a breadth of goals.
This problem, known as goal-conditioned reinforcement learning (GCRL), becomes especially challenging for long-horizon goals.
Current methods have tackled this problem by augmenting goal-conditioned policies with graph-based planning algorithms.
However, they struggle to scale to large, high-dimensional state spaces and assume access to exploration mechanisms for efficiently collecting training data.
In this work, we introduce Successor Feature Landmarks (\slm), a framework for exploring large, high-dimensional environments so as to obtain a policy that is proficient for any goal.
\slm leverages the ability of successor features (SF) to capture transition dynamics, using it to drive exploration by estimating state-novelty and to enable high-level planning by abstracting the state-space as a non-parametric landmark-based graph.
We further exploit SF to directly compute a goal-conditioned policy for inter-landmark traversal, which we use to execute plans to “frontier” landmarks at the edge of the explored state space.
We show in our experiments on MiniGrid and ViZDoom that \slm enables efficient exploration of large, high-dimensional state spaces and outperforms state-of-the-art baselines on long-horizon GCRL tasks\footnote{The demo video and code can be found at \url{https://2016choang.github.io/sfl}.}.

\end{abstract}

\cutsectionup
\section{Introduction}
\label{sec:introduction}
\cutsectiondown
Consider deploying a self-driving car to a new city.
To be practical, the car should be able to explore the city such that it can learn to traverse from any starting location to any destination, since the destination may vary depending on the passenger.
In the context of reinforcement learning (RL), this problem is known as goal-conditioned RL (GCRL)~\cite{Kaelbling1993gcrl, Kaelbling:1993:hierarchicallearning}.
Previous works~\cite{Schaul:ICML2015:UVFA, Andrychowivz:NIPS2017:HER, Nair:NeurIPS2018:RIG, pong:ICLR2018:TDM, levy:ICLR2019:HAC} have tackled this problem by learning a goal-conditioned policy (or value function) applicable to any reward function or ``goal.’’
However, the goal-conditioned policy often fails to scale to long-horizon goals~\cite{Huang:NIPS2019:Mapping} since the space of state-goal pairs grows intractably large over the horizon of the goal.

\begin{figure*}[t]
    \centering
    \vspace{20pt}
    \includegraphics[width=0.9\linewidth]{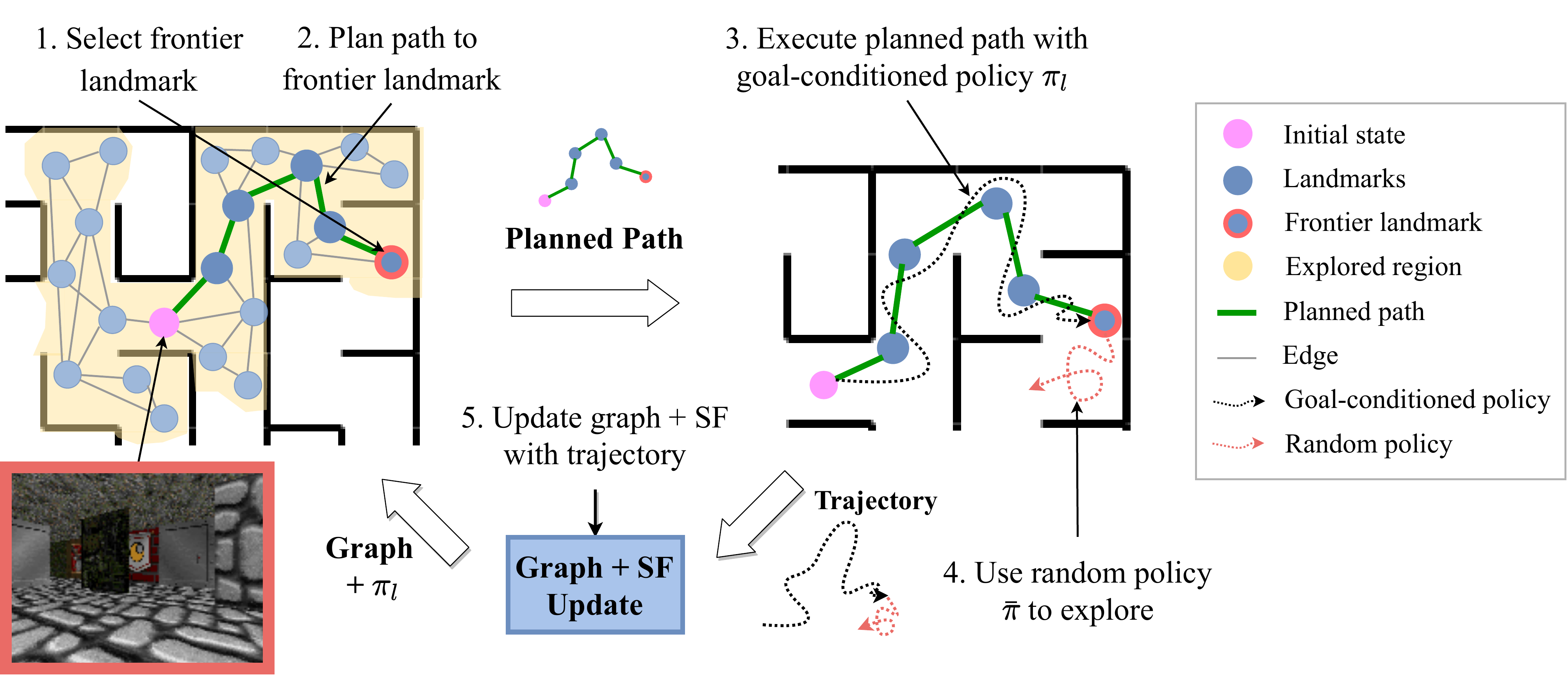}
    \caption{High-level overview of \slm.
\textbf{1.} During \textit{exploration}, select a frontier landmark (red circled dot) lying at the edge of the explored region as the target goal.
During \textit{evaluation} (not shown in the figure), the actual goal is selected as the target goal. 
\textbf{2.} Use the graph to plan a landmark path (green lines) to the target goal.
\textbf{3.} Execute the planned path with the goal-conditioned policy (black dotted arrow). 
\textbf{4.} During \textit{exploration}, upon reaching the frontier landmark, deploy the random policy (red dotted arrow) to reach novel states in unexplored areas.
\textbf{5.} Use each transition in the trajectory to update the graph and SF (see \Cref{fig:sfs-usage}). 
\textbf{Note:} The agent is never shown the top-down view of the maze and only uses first-person image observations (example on left) to carry out these steps.
Goals are also given as first-person images.}

    \label{fig:framework}
    \vspace*{-0.3in}
\end{figure*}

To address this challenge, the agent needs to (a) explore the state-goal space such that it is proficient for any state-goal pair it might observe during test time and (b) reduce the effective goal horizon for the policy learning to be tractable.
Recent work~\cite{nasiriany:NIPS2019:LEAP, Huang:NIPS2019:Mapping} has tackled long-horizon GCRL by leveraging model-based approaches to form plans consisting of lower temporal-resolution subgoals.
The policy is then only required to operate for short horizons between these subgoals.
One line of work learned a universal value function approximator (UVFA)~\cite{Schaul:ICML2015:UVFA} to make local policy decisions and to estimate distances used for building a landmark-based map, but assumed a low-dimensional state space where the proximity between the state and goal could be computed by the Euclidean distance~\cite{Huang:NIPS2019:Mapping}.
Another line of research focused on visual navigation tasks conducted planning over graph representations of the environment~\cite{Savinov:ICLR2018:SPTM, Eysenbach:NIPS2019:SoRB, Laskin:2020:SGM, chaplot:CVPR2020:neuralslam}.
However, these studies largely ignored the inherent exploration challenge present for large state spaces, and either assumed the availability of human demonstrations of exploring the state space~\cite{Savinov:ICLR2018:SPTM}, the ability to spawn uniformly over the state space~\cite{Eysenbach:NIPS2019:SoRB, Laskin:2020:SGM}, or the availability of ground-truth map information~\cite{chaplot:CVPR2020:neuralslam}.
In this work, we aim to learn an agent that can tackle long-horizon GCRL tasks and address the associated challenges in exploration.
Our key idea is to use successor features (SF)~\cite{Kulkarni:2016:DSRL, Barreto:NIPS2017:SuccessorFeatures} --- a representation that captures transition dynamics --- to define a novel distance metric, Successor Feature Similarity (SFS).
First, we exploit the transfer ability of SF to formulate a goal-conditioned value function in terms of SFS between the current state and goal state.
By just learning SF via self-supervised representation learning, we can directly obtain a goal-conditioned policy from SFS without any additional policy learning.
Second, we leverage SFS to build a landmark-based graph representation of the environment;
the agent adds observed states as landmarks based on their SFS-predicted novelty and forms edges between landmarks by using SFS as a distance estimate.
SF as an abstraction of transition dynamics is a natural solution for building this graph when we consider the MDP as a directed graph of states (nodes) and transitions (edges) following~\cite{Huang:NIPS2019:Mapping}.
We use this graph to systematically explore the environment by planning paths towards landmarks at the ``frontier'' of the explored state space and executing each segment of these planned paths with the goal-conditioned policy.
In evaluation, we similarly plan and execute paths towards (long-horizon) goals.
We call this framework \textit{Successor Feature Landmarks} (\slm), illustrated in \Cref{fig:framework}.

Our contributions are as follows:
(i) We use a single self-supervised learning component that captures dynamics information, SF, to build all the components of a graph-based planning framework, \slm.
(ii) We claim that this construction enables knowledge sharing between each module of the framework and stabilizes the overall learning. 
(iii) We introduce the SFS metric, which serves as a distance estimate and enables the computation of a goal-conditioned Q-value function \textit{without further learning}.
We evaluate \slm against current graph-based methods in long-horizon goal-reaching RL and visual navigation on \minigrid{}~\cite{Chevalier-Boisvert:2018:Minigrid}, a 2D gridworld, and \vizdoom{}~\cite{Wydmuch:2018:Vizdoom}, a visual 3D first-person view environment with large mazes.
We observe that \slm outperforms state-of-the-art navigation baselines, most notably when goals are furthest away.
In a setting where exploration is needed to collect training experience, \slm significantly outperforms the other methods which struggle to scale in \vizdoom{}'s high-dimensional state space.

\cutsectionup
\section{Related Work}
\label{sec:related_work}
\cutsectiondown

\textbf{Goal-conditioned RL}.
Prior work has tackled GCRL by proposing variants of goal-conditioned value functions such as UVFAs which estimate cumulative reward for any given state-goal pair~\cite{Moore:IJCAI1999:multivalue, Sutton:AAMAS2011:GVF, Schaul:ICML2015:UVFA, pong:ICLR2018:TDM}.
HER~\cite{Andrychowivz:NIPS2017:HER} improved the sample efficiency in training UVFAs by relabeling reached states as goals.
Mapping State Space (MSS)~\cite{Huang:NIPS2019:Mapping} then extended UVFAs to long-horizon tasks by using a UVFA as both a goal-conditioned policy and a distance metric to build a graph for high-level planning.
MSS also addressed exploration by selecting graph nodes to be at edge of the map's explored region via farthest point sampling.
However, this method was only evaluated in low-dimensional state spaces.
LEAP~\cite{nasiriany:NIPS2019:LEAP} used goal-conditioned value functions to form and execute plans over latent subgoals, but largely ignored the exploration question.
Conversely, other works~\cite{Pong:ICML2020:SkewFit, Chen:ICLR2019:Learning} have worked on exploration for goal-reaching policies, but do not tackle the long-horizon case.
ARC~\cite{Ghosh:ICLR2019:ARC} proposed learning representations that measure state similarity according to the output of a maximum entropy goal-conditioned policy, which can be utilized towards exploration and long-horizon hierarchical RL.
However, ARC assumes access to the goal-conditioned policy, which can be difficult to obtain in large-scale environments.
Our method can achieve both efficient exploration and long-horizon goal-reaching in high-dimensional state spaces with a SF-based metric that acts as a goal-conditioned value function and distance estimate for graph-building.

\textbf{Graph-based planning.}
Recent approaches have tackled long-horizon tasks, often in the context of visual navigation, by conducting planning on high-level graph representations and deploying a low-level controller to locally move between nodes; our framework also falls under this paradigm of graph-based planning.
Works such as SPTM~\cite{Savinov:ICLR2018:SPTM} and SoRB~\cite{Eysenbach:NIPS2019:SoRB} used a deep network as a distance metric for finding shortest paths on the graph, but rely on human demonstrations or sampling from the replay buffer to populate graph nodes.
SGM~\cite{Laskin:2020:SGM} introduced a two-way consistency check to promote sparsity in the graph, allowing these methods to scale to larger maps.
However, these methods rely on assumptions about exploration, allowing the agent to spawn uniformly random across the state space during training.
To address this, HTM~\cite{Liu:ICML2020:HTM} used a generative model to hallucinate samples for building the graph in a zero-shot manner, but was not evaluated in 3D visual environments.
NTS~\cite{chaplot:CVPR2020:neuralslam} achieved exploration and long-horizon navigation as well as generalization to unseen maps by learning a geometric-based graph representation, but required access to a ground-truth map to train their supervised learning model.
In contrast, our method achieves exploration by planning and executing paths towards landmarks near novel areas during training.
\slm also does not require any ground-truth data; it only needs to learn SF in a self-supervised manner.

\textbf{Successor features.}
Our work is inspired by recent efforts in developing successor features (SF)~\cite{Kulkarni:2016:DSRL, Barreto:NIPS2017:SuccessorFeatures}.
They have used SF to decompose the Q-value function into SF and the reward which enables efficient policy transfer~\cite{Barreto:NIPS2017:SuccessorFeatures, Borsa:ICLR2019:USFA} and to design transition-based intrinsic rewards~\cite{Machado:2018:CountExpSR, Zhang:2019:SID} for efficient exploration.
SF has also been used in the options framework~\cite{Sutton:1999:Options};
Eigenoptions~\cite{Machado:ICLR2018:EigenOptionSR} derived options from the eigendecomposition of SF, but did not yet apply them towards reward maximization.
Successor Options~\cite{Ramesh:IJCAI2019:SuccessorOptions} used SF to discover landmark states and design a latent reward for learning option policies, but was limited to low-dimensional state spaces.
Our framework leverages a SF-based similarity metric to formulate a goal-conditioned policy, abstract the state space as a landmark graph for long-horizon planning and to model state-novelty for driving exploration.
While the options policies proposed in these works have to be learned from a reward signal, we can obtain our goal-conditioned policy directly from the SF similarity metric without any policy learning.
To our knowledge, this is first work that uses SF for graph-based planning and long-horizon GCRL tasks.

\cutsectionup
\section{Preliminaries}
\label{sec:background}
\cutsectiondown

\def\goal{\mathcal{G}\xspace}
\subsection{Goal-Conditioned RL}
\cutsubsectiondown
\label{sec:problem}

Goal-conditioned reinforcement learning (GCRL) tasks~\cite{Kaelbling1993gcrl} are Markov Decision Processes (MDP) extended with a set of goals $\goal$ and defined by a tuple $(\mathcal{S}, \mathcal{A}, \goal, \mathcal{R}_{\goal}, \mathcal{T}, \gamma)$, where $\mathcal{S}$ is a state space, $\mathcal{A}$ an action set, $\mathcal{R}_{\goal}:\mathcal{S}\times\mathcal{A}\times\goal\rightarrow\mathbb{R}$ a goal-conditioned reward function, $\mathcal{T}$ the transition dynamics, and $\gamma$ a discount factor.
Following~\cite{Huang:NIPS2019:Mapping, venkattaramanujam2019self}, we focus on the setting where the goal space $\goal$ is a subset of the state space $\mathcal{S}$, and the agent can receive non-trivial rewards only when it can reach the goal (\ie, sparse-reward setting).
We aim to find an optimal goal-conditioned policy $\pi: \mathcal{S}\times \goal \rightarrow \mathcal{A}$ to maximize the expected cumulative reward, $V_{g}^{\pi}(s_0)=\mathbb{E}^{\pi}{ \left[ \sum_{t} \gamma^t r_t \right] }$; \ie, goal-conditioned value function.
We are especially interested in long-horizon tasks where goals are distant from the agent's starting state, requiring the policy to operate over longer temporal sequences.
\subsection{Successor Features}
\label{sec:SR}
\cutsectiondown

In the tabular setting, the successor representation (SR)~\cite{Dayan:1993:SR,Kulkarni:2016:DSRL} is defined as the expected discounted occupancy of futures state $s'$ starting in state $s$ and action $a$ and acting under a policy $\pi$:
\begin{equation}
    \label{eq: SR}
    M_\pi(s, a, s')= \mathbb{E}^{\pi}\left[ \textstyle \sum_{t'=t}^{\infty} {\gamma^{t' - t}\mathbb{I}(S_{t'} = s')} \middle\vert %
    S_t = s, A_t = a
    \right]
\end{equation}

The SR $M(s, a)$ is then a concatenation of $M(s, a, s'), \forall s \in S$. %
We may view SR as a representation of state similarity extended over the time dimension, as described in~\cite{Machado:ICLR2018:EigenOptionSR}.
In addition, we note that the SR is solely determined by $\pi$ and the transition dynamics of the environment $p (s_{t+1} | s_t, a_t)$.

Successor features (SF)~\cite{Barreto:NIPS2017:SuccessorFeatures, Kulkarni:2016:DSRL} extend SR~\cite{Dayan:1993:SR} to high-dimensional, continuous state spaces in which function approximation is often used.
SF's formulation modifies the definition of SR by replacing enumeration over all states $s'$ with feature vector $\phi_{s'}$.
The SF $\sf$ of a state-action pair $(s, a)$ is then defined as:
\begin{equation}
    \label{eq: SF}
    \sf(s, a) = \mathbb{E}^{\pi}\left[\textstyle \sum_{t'=t}^{\infty} {\gamma^{t' - t}\phi_{s_{t'}}} \,\middle\vert\, %
    S_t = s, A_t = a
    \right]
\end{equation}
In addition, SF can be defined in terms of only the state, %
$
    \sf(s) = \mathbb{E}_{a \sim \pi(s)}[\sf(s, a)].
$
SF allows decoupling the value function into the successor feature (dynamics-relevant information) with task (reward function):
if we assume that the one-step reward of transition $(s, a, s')$ with feature $\phi(s, a, s')$ can be written as $r(s, a, s') = \phi(s, a, s')^\top \mathbf{w}$, where $\mathbf{w}$ are learnable weights to fit the reward function, we can write the Q-value function as follows \cite{Barreto:NIPS2017:SuccessorFeatures, Kulkarni:2016:DSRL}:
\begin{align}
        Q^{\pi}(s, a)  &= \mathbb{E}^{\pi}\left[\sum_{t'=t}^{\infty}{\gamma^{t'-t}r(S_{t'}, A_{t'}, S_{t'+1}) \,\Big\vert\,     %
        S_t = s, A_t = a
        }
        \right] \nonumber \\
                    &= \mathbb{E}^{\pi}\left[\sum_{t'=t}^{\infty} \gamma^{t'-t}\phi_{t'}^{\top} \mathbf{w} \,\Big\vert\, %
                    S_t = s, A_t = a
                    \right]  %
                    = \sf(s, a)^\top \mathbf{w} \label{eq: SF-Q}
\end{align}
Consequently, the Q-value function separates into SF, which represents the policy-dependent transition dynamics, and the reward vector $\bf{w}$ for a particular task. 
Later in \Cref{sec:local-policy}, we will extend this formulation to the goal-conditioned setting and discuss our choices for $\mathbf{w}$.

\cutsectionup
\section{Successor Feature Landmarks}
\label{sec:methods}
\cutsectiondown

\begin{algorithm}[t]
	\caption{\textsf{ Training}}
	\label{alg:overall2}
	\begin{algorithmic}[1]
		\STATE \textbf{Initialize:} Graph $G=(L, E)$, parameter $\theta$ of $\sfs_{\theta}$, replay buffer $D$, hyperparameter $T_\text{exp}$, landmark transition count $N^l$
		\WHILE{env not done}
		\STATE $\landfront\sim \mathrm{Softmax}(\frac{1}{\mathrm{Count}(L)})$\hfill\COMMENT{Choose frontier landmark via count-based sampling}
		\WHILE{$\landcurr \neq \landfront$}
			\STATE $\landtarget \gets \text{PLAN}(G, \landcurr, \landfront)$\hfill\COMMENT{Plan path to frontier landmark}
			\STATE $\tau_\text{traverse}, \landcurr \gets \textsf{Traverse}(\pi_l, \landtarget)$\hfill\COMMENT{Traverse to $\landtarget$ with $\pi_l$ (\Cref{alg:traverse} in~\S\ref{sec:appendix-algorithm})}
			\STATE $G, N^l\gets \textsf{Graph-Update}(G, \tau_\text{traverse}, \sfs_{\theta}, N^l)$
			\hfill\COMMENT{Update graph (\Cref{alg:landmark-graph})}
		\ENDWHILE
		\STATE $\tau_\text{rand}=\{s_t, a_t, r_t\}_{t}^{T_\text{exp}}\sim\bar{\pi}$\hfill\COMMENT{Explore with random policy for $T_\text{exp}$ steps}
		\STATE $G, N^l\gets \textsf{Graph-Update}(G, \tau_\text{rand}, \sfs_{\theta}, N^l$)\hfill\COMMENT{Update graph (\Cref{alg:landmark-graph})}
		\STATE $D\gets D\cup \tau_\text{random}$
		\STATE Update $\theta$ from TD error with mini-batches sampled from $D$\hfill\COMMENT{Update SF parameters}
		\ENDWHILE
	\end{algorithmic}
\end{algorithm} %

\begin{figure*}[t]
    \centering
    \includegraphics[width=\linewidth]{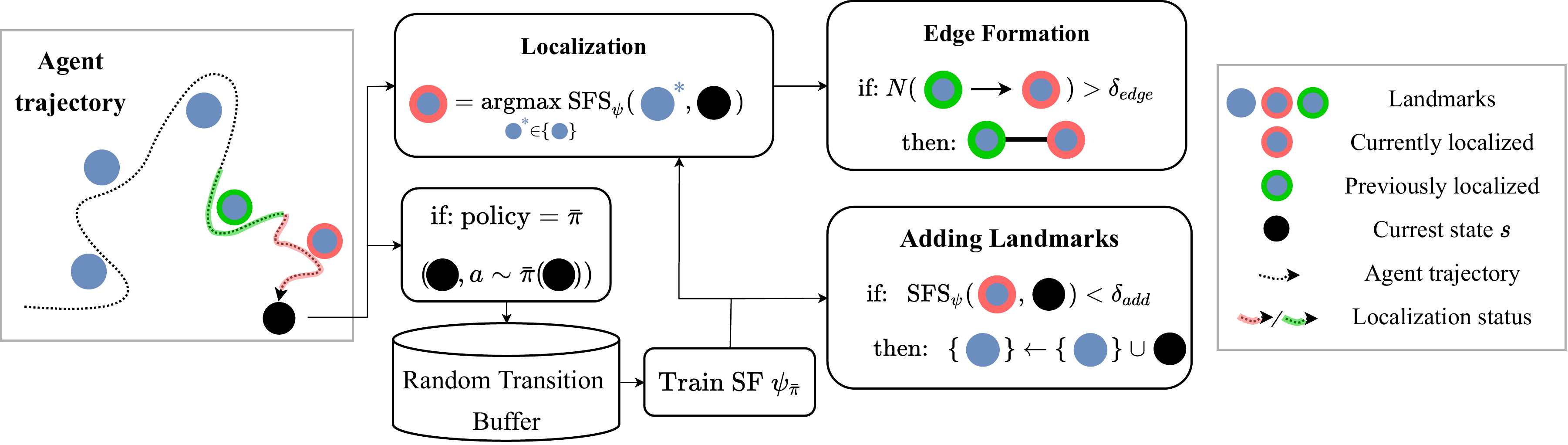}
    \caption{The \textbf{Graph + SF Update} step occurs after every transition $(s, a)$. 
The agent computes SFS between current state $s$ (black dot) and all landmarks (blue dots). 
It then localizes itself to the nearest landmark (red circled dot). 
The agent records transitions between the previously localized landmark (green circled dot) and this landmark.
If the number of transitions between two landmarks is greater than $\delta_{edge}$, then an edge is formed between them.
If SFS between the current localized landmark
and $s$ is less than $\delta_{add}$, then $s$ is added as a landmark.
Finally, transitions generated by the random policy are added to the
random transition buffer, and SF is trained on batch samples from this buffer.}
    \label{fig:sfs-usage}
    \vspace*{-0.1in}
\end{figure*}

We present Successor Feature Landmarks (\slm), a graph-based planning framework for supporting exploration and long-horizon GCRL.
\slm is centrally built upon our novel distance metric: Successor Feature Similarity (SFS, \S  \ref{sec:sfs}).
We maintain a non-parametric graph of state ``landmarks,'' using SFS as a distance metric for determining which observed states to add as landmarks and how these landmarks should be connected (\S \ref{sec:landmark-graph}).
To enable traversal between landmarks, we directly obtain a local goal-conditioned policy from SFS between current state and the given landmark (\S \ref{sec:local-policy}).
With these components, we tackle the long-horizon setting by planning on the landmark graph and finding the shortest path to the given goal, which decomposes the long-horizon problem into a sequence of short-horizon tasks that the local policy can then more reliably achieve (\S \ref{sec:planning}).
In training, our agent focuses on exploration.
We set the goals as ``frontier'' landmarks lying at the edge of the explored region, and use the planner and local policy to reach the frontier landmark. 
Upon reaching the frontier, the agent locally explores with a random policy and uses this new data to update its SF and landmark graph (\S \ref{sec:exploration}).
In evaluation, we add the given goal to the graph and follow the shortest path to it.
Figure~\ref{fig:framework} illustrates the overarching framework and Figure~\ref{fig:sfs-usage} gives further detail into how the graph and SF are updated.
Algorithm~\ref{alg:overall2} describes the procedure used to train \slm.

\cutsubsectionup
\subsection{Successor Feature Similarity} \label{sec:sfs}
\cutsubsectiondown

SFS, the foundation of our framework, is based on SF.
For context, we estimate SF $\psi$ as the output of a deep neural network parameterized by $\theta: \sf(s, a) \approx \sf_{\theta}(\phi(s), a)$, where $\phi$ is a feature embedding of state and $\pi$ is a fixed policy which we choose to be a uniform random policy denoted as $\bar{\pi}$.
We update $\theta$ by minimizing the temporal difference (TD) error~\cite{Kulkarni:2016:DSRL, Barreto:NIPS2017:SuccessorFeatures}. %
Details on learning SF are provided in \Cref{sec:appendix-sf-learning}.

Next, to gain intuition for SFS, suppose we wish to compare two state-action pairs $(s_1, a_1)$ and $(s_2, a_2)$ in terms of similarity.
One option is to compare $s_1$ and $s_2$ directly via some metric such as $\ell_2$ distance, but this ignores $a_1, a_2$ and dynamics of the environment.

To address this issue, we should also consider the states the agent is expected to visit when starting from each state-action pair, for a fixed policy $\pi$.
We choose $\pi$ to be uniform random, i.e. $\bar{\pi}$, so that only the dynamics of the environment will dictate which states the agent will visit.
With this idea, we can define a novel similarity metric, \textbf{Successor Feature Similarity (SFS)}, which measures the similarity of the expected discounted state-occupancy of two state-action pairs.
Using the successor representation $M_{\bar{\pi}}(s, a)$ as defined in~\Cref{sec:SR}, we can simply define SFS as the dot-product between the two successor representations for each state-action pair:
\begin{align}
    \begin{split}
    &\sfs ((s_1, a_1), (s_2, a_2)) \\
    & %
    = \sum_{s' \in S}
        \mathbb{E}^{\bar{\pi}}\left[\sum_{t'=t}^{\infty} {\gamma^{t' - t}\mathbb{I}(S_t = s')} \,\middle\vert\, \stackanchor{$S_t = s_1$,}{$A_t = a_1$} \right] %
        \times \mathbb{E}^{\bar{\pi}}\left[\sum_{t'=t}^{\infty} {\gamma^{t' - t}\mathbb{I}(S_t = s')} \,\middle\vert\, \stackanchor{$S_t = s_2$,}{$A_t = a_2$} \right] \\
    & 
    = \sum_{s' \in S} M_{\bar{\pi}}(s_1, a_1, s') \times M_{\bar{\pi}}(s_2, a_2, s') = M_{\bar{\pi}}(s_1, a_1)^\top M_{\bar{\pi}}(s_2, a_2)
    \end{split}
\end{align}

We can extend SFS to the high-dimensional case by encoding states in the feature space $\phi$ and replacing $M_{\bar{\pi}}(s, a)$ with $\barsf(s, a)$.
The intuition remains the same, but we instead measure similarities in the feature space.
In practice, we normalize $\sf(s, a)$ before computing SFS to prevent high-value feature dimensions from dominating the similarity metric, hence defining SFS as the cosine similarity between SF.
In addition, we may define SFS between just two states by getting rid of the action dimension in SF: %
\begin{align}
    \label{eq:SSF}
    \sfs((s_1, a_1), (s_2, a_2)) &= \barsf(s_1, a_1)^\top \barsf(s_2, a_2)
    \\
    \label{eq:SSF-ss}
    \sfs(s_1, s_2) &= \barsf(s_1)^\top \barsf(s_2)
\end{align}

\cutsubsectionup
\subsection{Landmark Graph}\label{sec:landmark-graph}
\cutsubsectiondown
The landmark graph $G$ serves as a compact representation of the state space and its transition dynamics.
$G$ is dynamically-populated in an online fashion as the agent explores more of its environment.
Formally, landmark graph $G=(L, E)$ is a tuple of landmarks $L$ and edges $E$.
The landmark set $L=\{l_1, \ldots, l_{|L|}\}$ is a set of states representing the explored part of the state-space.
The edge set $E$ is a matrix $\mathbb{R}^{|L|\times |L|}$, where $E_{i, j}=1$ if $l_i$ and $l_j$ is connected, and 0 otherwise.
Algorithm~\ref{alg:landmark-graph} outlines the graph update process, and the following paragraph describes this process in further detail.
\cutparagraphup
\paragraph{Agent Localization and Adding Landmarks} \label{para:adding-localization}
At every time step $t$, we compute the landmark closest to the agent under SFS metric: $\landcurr = \argmax_{l\in L}\sfs(s_t, l)$.
If $\sfs(s_t, \landcurr) < \delta_{add}$, the add threshold, then we add $s_t$ to the landmark set $L$. 
Otherwise, if $\sfs(s_t, \landcurr) > \delta_{\text{local}}$, the localization threshold, then the agent is localized to $\landcurr$.
If the agent was previously localized to a different landmark $\landprev$, then we increment the count of the landmark transition \smash{$N^l_{(\landprev \rightarrow \landcurr)}$} by 1 where $N^l\in\mathbb{N}^{|E|}$ is the landmark transition count, which is used to form the graph edges.
\footnote{Zhang et al.~\cite{Zhang:ICML2018:Composable} proposed a similar idea of recording the transitions between sets of user-defined attributes.
We extend this idea to the function approximation setting where landmark attributes are their SFs.}

Since we progressively build the landmark set, we maintain all previously added landmarks.
As described above, this enables us to utilize useful landmark metrics such as how many times the agent has been localized to each landmark and what transitions have occurred between landmarks to improve the connectivity quality of the graph.
In comparison, landmarks identified through clustering schemes such as in Successor Options \cite{Ramesh:IJCAI2019:SuccessorOptions} cannot be used in this manner because the landmark set is rebuilt every few iterations.
See \Cref{sec:appendix-landmark-formation} for a detailed comparison on landmark formation.

\cutparagraphup
\paragraph{Edge Formation}
\label{para:edge-formation} 
\begin{wrapfigure}{r}{0.53\textwidth}
    \vspace*{-13pt}
    \begin{minipage}{0.53\textwidth}
        \begin{algorithm}[H]
        \caption{\textsf{Graph-Update ({\S \ref{sec:landmark-graph}})}}
            \label{alg:landmark-graph}
            \begin{algorithmic}[1]
                \INPUT{Graph $G=(L, E)$, $\sfs_{\theta}$, trajectory $\tau$,
                \\ landmark transition count $N^l$}
                \OUTPUT{updated graph $G$ and $N^l$}
                \STATE $\landprev \gets \emptyset$\hfill\COMMENT{Previously localized landmark}
                \FOR{$s \in \tau$}
                    \STATE $\landcurr\gets \argmax_{l\in L}\sfs_{\theta}(s, l)$ \hfill\COMMENT{Localize}
                    \IF{$\sfs_{\theta}(s, \landcurr) < \delta_{add}$}
                        \STATE $L \gets L\cup s$\hfill\COMMENT{Add landmark}
                    \ENDIF
                    \IF{$\sfs_{\theta}(s, \landcurr) > \delta_{local}$}
                        \IF{$\landprev \neq \emptyset$ \AND $\landprev \neq \landcurr $}
                            \STATE $N^l_{(\landprev\rightarrow \landcurr)}\gets N^l_{(\landprev\rightarrow \landcurr)}+1$\hfill\COMMENT{Record landmark transition}
                            \IF{$N^l_{(\landprev\rightarrow \landcurr)} > \delta_{edge}$}
                                \STATE $E \gets E\cup (\landprev \rightarrow \landcurr)$\hfill\COMMENT{Form edge}
                            \ENDIF
                        \ENDIF
                        \STATE $\landprev \gets \landcurr$
                    \ENDIF
                \ENDFOR

                \STATE \textbf{return} $G,\ N^l$
            \end{algorithmic}
        \smallskip
        \end{algorithm}
    \end{minipage}
    \vspace{-20pt}
\end{wrapfigure} %

We form edge $E_{i, j}$ if the number of the landmark transitions is larger than the edge threshold, \ie, $N^l_{l_i\rightarrow l_j} > \delta_{edge}$, with weight $\smash{W_{i, j} = \exp({-(N^l_{l_i\rightarrow l_j})}})$.
We apply filtering improvements to $E$ in \vizdoom{} to mitigate the perceptual aliasing problem where faraway states can appear visually similar due to repeated use of textures.
See \Cref{sec:appendix-aliasing} for more details.

\subsection{Local Goal-Conditioned Policy}
\label{sec:local-policy}
\cutsubsectiondown
We want to learn a local goal-conditioned policy $\localpi: \mathcal S \times \mathcal G \rightarrow \mathcal A$ to reach or transition between landmarks.
To accomplish this, $\localpi$ should maximize the expected return $V(s, g) = \mathbb{E} [\sum_{t=0}^{\infty} \gamma^t r(s_t, a_t, g)]$, where $r(s, a, g)$ is a reward function that captures how close $s$ (or more precisely $s, a$) is to the goal $g$ in terms of feature similarity:
\begin{align}
    \label{eq:reward_sf_goalpolicy}
    r(s, a, g) = \phi(s, a)^\top \randsf(g),
\end{align}
where $\randsf$ is the SF with respect to the random policy $\randpi$. 
Recall that we can decouple the Q-value function into the SF representation and reward weights $\bf{w}$ as shown in Eq.~(\ref{eq: SF-Q}).
The reward function \Cref{eq:reward_sf_goalpolicy} is our deliberate choice rather than learning a linear reward regression model \cite{Barreto:NIPS2017:SuccessorFeatures}, so the value function can be instantly computed.
If we let $\bf{w}$ be $\randsf(g)$, we can have the Q-value function $Q^{\randpi}(s, a, g)$ for the goal-conditioned policy $\pi(a|s, g)$ being equal to the SFS between $s$ and $g$:
\begin{align}
    \label{eq:sfs q}
        Q^{\randpi}(s, a, g) &= \randsf(s, a)^\top \randsf(g) = \randsfs(s, a, g).
\end{align}

The goal-conditioned policy is derived by sampling actions from the goal-conditioned Q-value function in a greedy manner for discrete actions.
In the continuous action case, we can learn the goal-conditioned policy by using a compatible algorithm such as DDPG \cite{Lillicrap::ICLR2016::DDPG}.
However, extending SF learning to continuous action spaces is beyond the scope of this work and is left for future work.

\cutsubsectionup
\subsection{Planning}
\label{sec:planning}
\cutsubsectiondown
Given the landmark graph $G$, we can plan the shortest-path from the landmark closest to the agent $\landcurr$ to a final target landmark $\landtarget$ by selecting a sequence of landmarks $[l_0, l_1, \ldots, l_k]$ in the graph $G$ with minimal weight (see $\S$\ref{para:edge-formation}) sum along the path, where $l_0=\landcurr$, $l_k=\landtarget$, and $k$ is the length of the plan.
In training, we use frontier landmarks $\landfront$ which have been visited less frequently as $\landtarget$.
In evaluation, the given goal state is added to the graph and set as $\landtarget$.
See Algorithm~\ref{alg:overall2} for an overview of how planning is used to select the agent's low-level policy in training.

\cutsubsectionup
\subsection{Exploration} \label{sec:exploration}
\cutsubsectiondown
We sample frontier landmarks $\landfront$ proportional to the inverse of their visitation count (\ie, with count-based exploration).
We use two policies: a local policy $\localpi$ for traversing between landmarks and a random policy $\randpi$ for exploring around a frontier landmark.
Given a target frontier landmark $\landfront$, we construct a plan $[l_0, l_1, \ldots, \landfront]$.
When the agent is localized to landmark $l_{i}$, the policy at time $t$ is defined as $\localpi(a|s_t; l_{i+1})$.
When the agent is localized to a landmark that is not included in the current plan $[l_0, l_1, \ldots, \landfront]$, then it re-plans a new path to $\landfront$.
Such failure cases of transition between the landmarks are used to prevent edge between those landmarks from being formed (see~\Cref{sec:appendix-aliasing} for details).
We run this process until either $\landfront$ is reached or until the step-limit  $N_{\text{front}}$ is reached.
At that point, random policy $\randpi$ is deployed for exploration for  $N_{\text{explore}}$ steps, adding novel states to our graph as a new landmark.
While the random policy is deployed, its trajectory $\tau\sim\randpi$ is added to the random transition buffer.
SF $\barsf_{\theta}$ is updated with batch samples from this buffer.

The random policy is only used to explore local neighborhoods at frontier regions for a short horizon while the goal-conditioned policy and planner are responsible for traveling to these frontier regions.
Under this framework, the agent is able to visit a diverse set of states in a relatively efficient manner.
Our experiments on large \vizdoom{} maps demonstrate that this strategy is sufficient for learning a SF representation that ultimately outperforms baseline methods on goal-reaching tasks.

\smallskip

\cutsectionup
\section{Experiments}
\label{sec:experiments}
\cutsectiondown

In our experiments, we evaluate the benefits of \slm for exploration and long-horizon GCRL. We first study how well our framework supports reaching long-horizon goals when the agent's start state is randomized across episodes. Afterwards, we consider how \slm performs when efficient exploration is required to reach distant areas by using a setting where the agent's start state is fixed in training.

\setlength\intextsep{0pt}
\begin{wrapfigure}{r}{0.45\textwidth}
    \centering
    \includegraphics[width=1\linewidth]{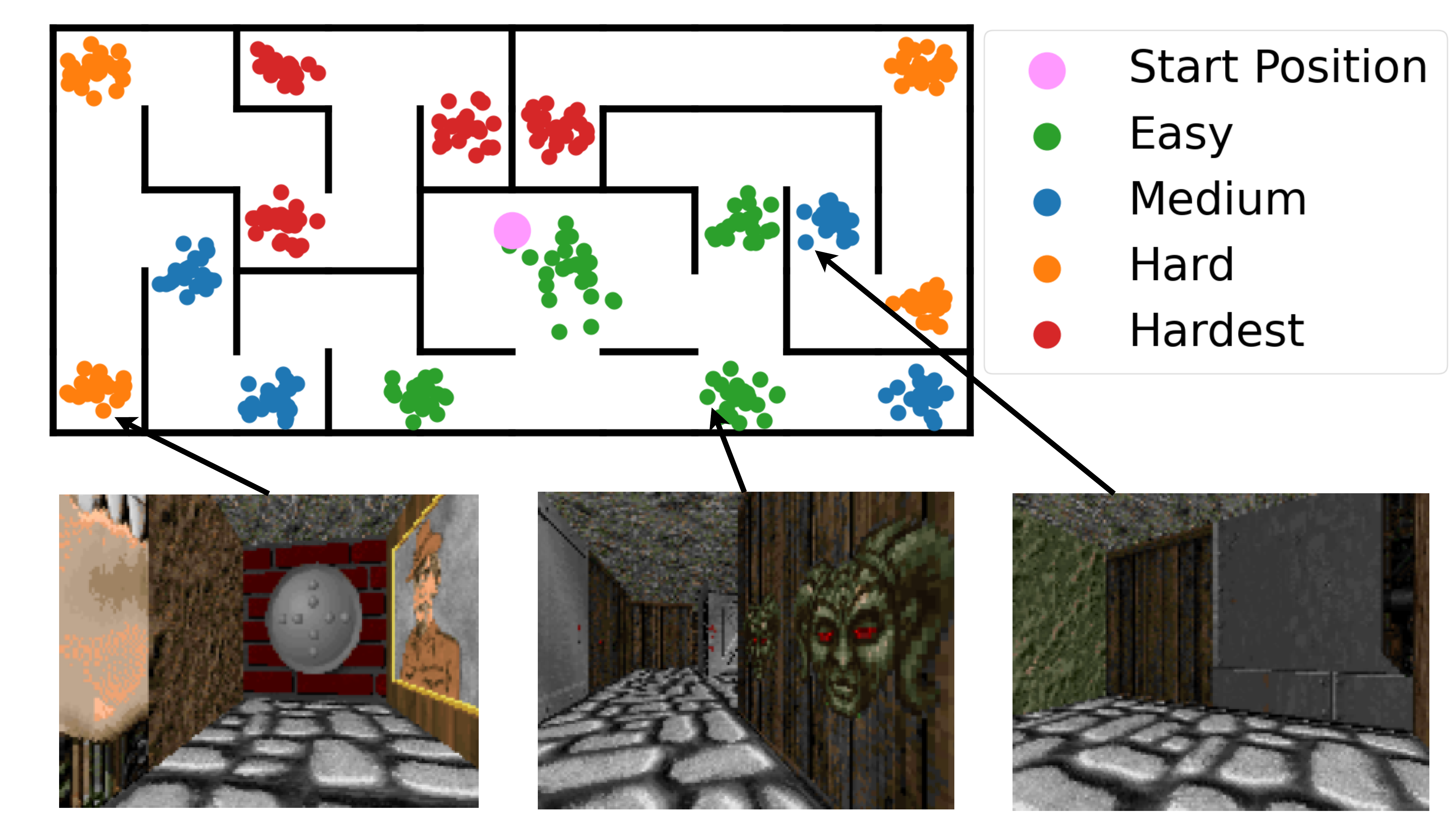}
    \caption{Top-down view of a \vizdoom{} maze used in \textit{fixed spawn} with sampled goal locations. 
    Examples of image goals given to the agent are shown at the bottom.
    Note that the agent cannot access the top-down view.
    }
    \label{fig:vizdoom}
\end{wrapfigure}

\subsection{Domain and Goal-Specification}
\label{sec:domains}
\cutsubsectiondown
\vizdoom{} is a visual navigation environment with 3D first-person observations. 
In \vizdoom{}, the large-scale of the maps and reuse of similar textures make it particularly difficult to learn distance metrics from the first-person observations. 
This is due to perceptual aliasing, where visually similar images can be geographically far apart.
We use mazes from SPTM in our experiments, with one example shown in Figure~\ref{fig:vizdoom}~\cite{Savinov:ICLR2018:SPTM}.

\minigrid{} is a 2D gridworld with tasks that require the agent to overcome different obstacles to access portions of the map.
We experiment on \fourroom{}, a basic 4-room map, and \multiroom{}, where the agent needs to open doors to reach new rooms.

We study two settings:
\begin{enumerate}[labelindent=0cm,labelsep=2pt,noitemsep,nolistsep, topsep=0pt,leftmargin=*]
    \item \textbf{Random spawn}: In training, the agent is randomly spawned across the map with no given goal. In evaluation, the agent is then tested on pairs of start and goal states, where goals are given as images. This enables us to study how well the landmark graph supports traversal between arbitrary start-goal pairs. 
    Following~\cite{Laskin:2020:SGM}, we evaluate on \texttt{easy}, \texttt{medium}, and \texttt{hard} tasks where the goal is sampled within 200m, 200-400m, and 400-600m from the initial state, respectively.
    \item \textbf{Fixed spawn}: In training, the agent is spawned at a fixed start state $s_\text{start}$ with no given goal. This enables us to study how well the agent can explore the map given a limited step budget per episode. In evaluation, the agent is again spawned at $s_\text{start}$ and is given different goal states to reach. In \vizdoom{}, we sample goals of varying difficulty accounting for the maze structure as shown in Figure~\ref{fig:vizdoom}. %
    In \minigrid{}, a similar setup is used except only one goal state is used for evaluation.
\end{enumerate}

\subsection{Baseline Methods for Comparison}
\label{sec:baselines}
\cutsubsectiondown
\textbf{Random spawn experiments}. We compare \slm against baselines used in SGM, as described below.
For each difficulty, we measure the average success rate over 5 random seeds.
We evaluate on the map used in SGM, \textit{SGM-Map}, and two more maps from SPTM, \textit{Test-2} and \textit{Test-6}.
\begin{enumerate}[labelindent=0cm,labelsep=3pt,noitemsep,nolistsep, topsep=0pt,leftmargin=*]
    \item \textbf{Random Actions}: random acting agent. Baseline shows task difficulty.
    \item \textbf{Visual Controller}: model-free visual controller learned via inverse dynamics. Baseline highlights how low-level controllers struggle to learn long-horizon policies and the benefits of planning to create high-level paths that the controller can follow.
    \item \textbf{SPTM}~\cite{Savinov:ICLR2018:SPTM}: planning module with a reachability network to learn a distance metric for localization and landmark graph formation. Baseline is used to measure how the SFS metric can improve localization and landmark graph formation.
    \item \textbf{SGM}~\cite{Laskin:2020:SGM}: data structure used to improve planning by inducing sparsity in landmark graphs. Baseline represents a recent landmark-based approach for long-horizon navigation.
\end{enumerate}

\textbf{Fixed spawn experiments}. In the \textit{fixed spawn} setting, we compare \slm against Mapping State Space (MSS)~\cite{Huang:NIPS2019:Mapping}, a UVFA and landmark-based approach for exploration and long-horizon goal-conditioned RL, as well as SPTM and SGM.
Again, we measure the average success rate over 5 random seeds.
We adapt the published code\footnote{\url{https://github.com/FangchenLiu/map_planner}} to work on \vizdoom{} and \minigrid{}.
To evaluate SPTM and SGM, we populate their graphs with exploration trajectories generated by Episodic Curiosity (EC)~\cite{Savinov:ICLR2019:EC}.
EC learns an exploration policy by using a reachability network to determine whether an observation is novel enough to be added to the memory and rewarding the agent every time one is added.
\Cref{sec:appendix-implementation-details} further discusses the implementation of these baselines.

\subsection{Implementation Details}
\label{sec:implementation-details}
\cutsubsectiondown

\slm is implemented with the \textit{rlpyt} codebase \cite{Stooke:2019:rlpyt}.
For experiments in \vizdoom{}, we use the pretrained ResNet-18 backbone from SPTM as a fixed feature encoder which is similarly used across all baselines.
For \minigrid{}, we train a convolutional feature encoder using time-contrastive metric learning~\cite{Sermanet:ICRA2017:TCN}.
Both feature encoders are trained in a self-supervised manner and aim to encode temporally close states as similar feature representations and temporally far states as dissimilar representations.
We then approximate SF with a fully-connected neural network, using these encoded features as input.
See \Cref{sec:appendix-implementation-details} for more details on feature learning, edge formation, and hyperparameters.
\subsection{Results}
\label{sec:results}
\cutsubsectiondown

\bgroup
    \def\arraystretch{1.1}
    \setlength{\tabcolsep}{4pt}
    \begin{table}[t]
        \footnotesize
        \begin{center}
            \begin{tabular}{ c | c c c | c c c | c c c } 
             \toprule
           Method & \multicolumn{3}{c|}{\textit{SGM-Map}} & \multicolumn{3}{c|}{\textit{Test-2}} & \multicolumn{3}{c}{\textit{Test-6}}\\
                & Easy & Medium & Hard & Easy & Medium & Hard & Easy & Medium & Hard \\ 
             \midrule %
             Random Actions & 58\% & 22\% & 12\% & 70\% & 39\% & 16\% & 80\% & 31\% & 18\% \\ 
             Visual Controller & 75\% & 35\% & 19\% & 83\% & 51\% & 30\% & 89\% & 39\% & 20\% \\ 
             SPTM \cite{Savinov:ICLR2018:SPTM} & 70\% & 34\% & 14\% & 78\% & 48\% & 18\% & 88\% & 40\% & 18\% \\ 
             SGM \cite{Laskin:2020:SGM} & \textbf{92}\% & 64\% & 26\% & \textbf{86\%} & 54\% & 32\% & 83\% & 43\% & 27\% \\ %
             \slm [Ours] & \textbf{92}\% & \textbf{82\%} & \textbf{67\%} & 82\% & \textbf{66\%} & \textbf{48\%} & \textbf{92\%} & \textbf{66\%} & \textbf{60\%} \\ %
             \bottomrule %
            \end{tabular}
        \vspace*{-0.1in}
        \caption{(\textit{Random spawn}) The success rates of compared methods on three \vizdoom{} maps.}
        \label{table:random-vizdoom}
        \end{center}
        \vspace*{-0.20in}
    \end{table}
\egroup

\textbf{Random Spawn Results}.
As shown in Table~\ref{table:random-vizdoom}, our method outperforms the other baselines on all settings.
\slm's performance on the Hard setting particularly illustrates its ability to reach long-horizon goals.
In terms of sample efficiency, \slm utilizes a total of 2M environment steps to simultaneously train SF and build the landmark graph.
For reference, SPTM and SGM train their reachability and low-level controller networks with over 250M environment steps of training data collected on \textit{SGM-Map}, with SGM using an additional 114K steps to build and cleanup their landmark graph.
For \textit{Test-2} and \textit{Test-6}, we fine-tune these two networks with 4M steps of training data collected from each new map to give a fair comparison.

\bgroup
    \def\arraystretch{1.1}
    \setlength{\tabcolsep}{4pt}
    \begin{table}[t]
        \footnotesize
        \begin{center}
            \begin{tabular}{ c | c c c c | c c c c } 
             \toprule
           Method & \multicolumn{4}{c|}{\textit{Test-1}} & \multicolumn{4}{c}{\textit{Test-4}}\\
                & Easy & Medium & Hard & Hardest & Easy & Medium & Hard & Hardest \\ 
             \midrule %
             MSS~\cite{Huang:NIPS2019:Mapping} & 23\% & 9\% & 1\% & 1\% & 21\% & 7\% & 7\% & 7\% \\
             EC~\cite{Savinov:ICLR2019:EC} + SPTM~\cite{Savinov:ICLR2018:SPTM} & 48\% & 16\% & 2\% & 0\% & 20\% & 10\% & 4\% & 0\%  \\ 
             EC~\cite{Savinov:ICLR2019:EC} + SGM~\cite{Laskin:2020:SGM} & 43\% & 3\% & 0\% & 0\% & 28\% & 7\% & 4\% & 1\%\\
             \slm [Ours] & \textbf{85}\% & \textbf{59\%} & \textbf{62\%} & \textbf{50\% }& \textbf{66\%} & \textbf{44\%} & \textbf{27\%} & \textbf{23\%} \\
             \bottomrule %
            \end{tabular}
        \vspace*{-0.1in}
        \caption{(\textit{Fixed spawn}) The success rates of compared methods on three \vizdoom{} maps.}
        \label{table:fixed-vizdoom}
        \end{center}
    \end{table}
\egroup

\begin{figure}[b]
\begin{minipage}[b]{0.49\textwidth}
    \centering
    \includegraphics[width=\linewidth]{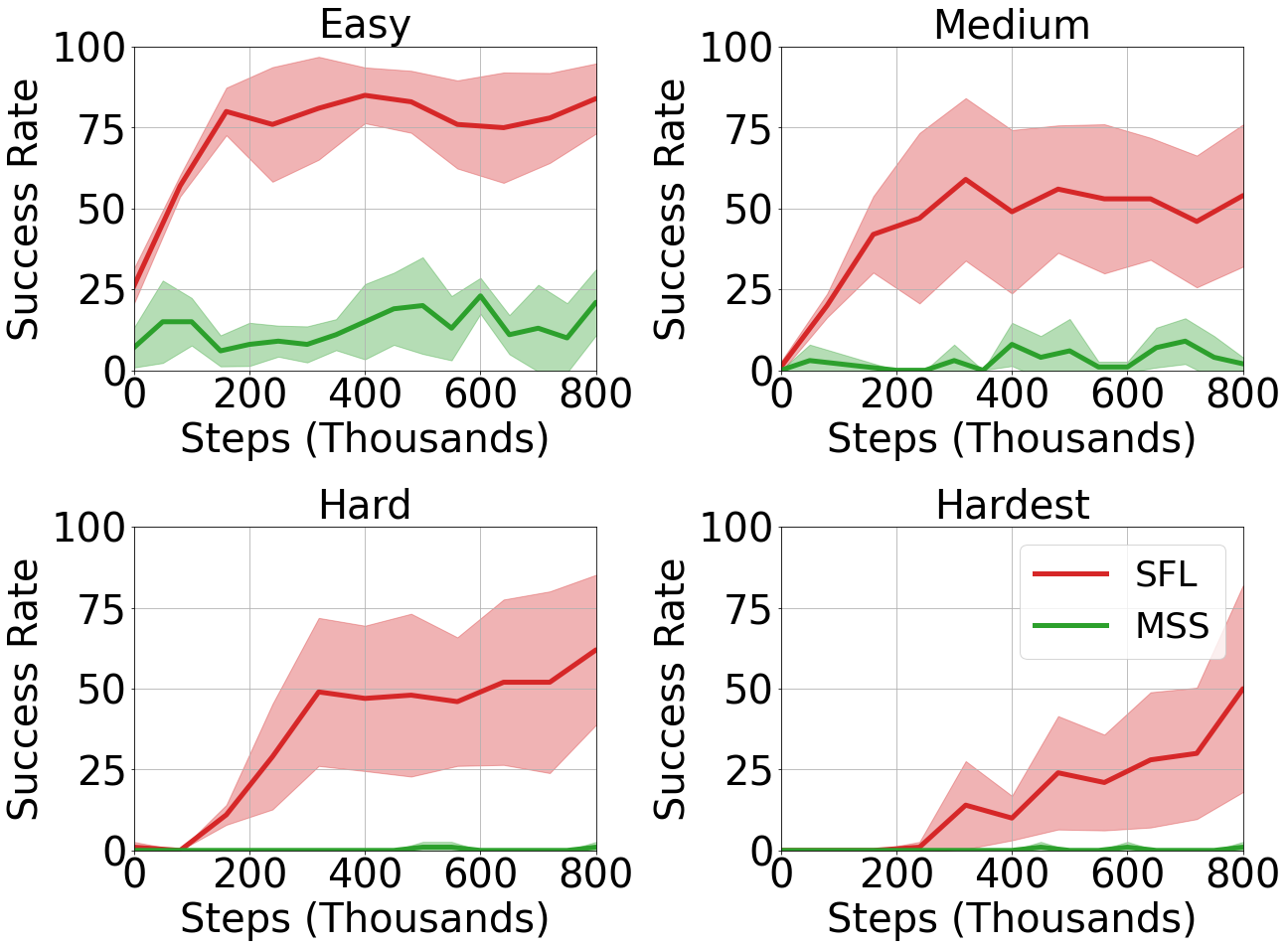}
    \caption{
    \textit{Fixed spawn} experiments on \vizdoom{} comparing \slm (red) to MSS (green) over number of environment steps for varying difficulty levels.}
    \label{fig:fixed-spawn-vizdoom-success-rate}
\end{minipage}
\hfill
\begin{minipage}[b]{0.49\textwidth}
    \centering
    \includegraphics[width=\linewidth]{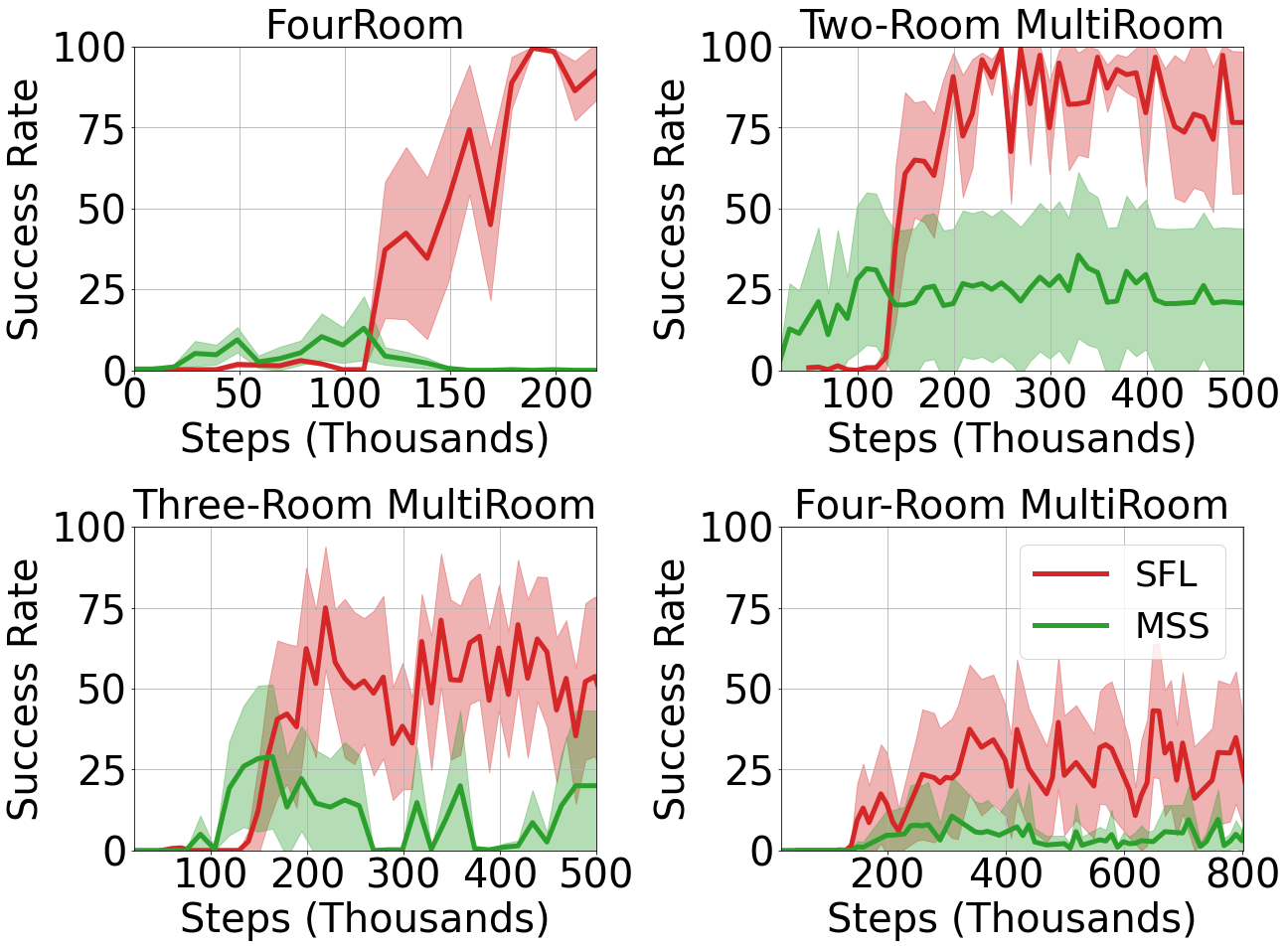}
    \caption{
    \textit{Fixed spawn} experiments on \minigrid{} comparing \slm (red) to MSS (green) over number of environment steps for varying difficulty levels.}
    \label{fig:fixed-spawn-minigrid-success-rate}
\end{minipage}

\end{figure}

\textbf{Fixed Spawn Results.}
We see in \Cref{table:fixed-vizdoom} that \slm reaches significantly higher success rates than the baselines across all difficulty levels, especially on \textit{Hard} and \textit{Hardest}.
\Cref{fig:fixed-spawn-vizdoom-success-rate} shows the average success rate over the number of environment steps for \slm (red) and MSS (green).
We hypothesize that MSS struggles because its UVFA is unable to capture geodesic distance in \vizdoom{}'s high-dimensional state space with first-person views.
The UVFA in MSS has to solve the difficult task of approximating the number of steps between two states, which we conjecture requires a larger sample complexity and more learning capacity.
In contrast, we only use SFS to relatively compare states, i.e. is SFS of state A higher than SFS of state B with respect to reference state C?
EC-augmented SPTM and SGM partially outperform MSS, but cannot scale to harder difficulty levels.
We suggest that these baselines suffer from disjointedness of exploration and planning: the EC exploration module is less effective because it does not utilize planning to efficient reach distant areas, which in turn limits the training of policy networks.
See~\Cref{sec:appendix-MSS-analysis} and~\Cref{sec:appendix-ec-analysis} for more analysis on the baselines.

Figure~\ref{fig:fixed-spawn-minigrid-success-rate} shows the average success rate on \minigrid{} environments, where \slm (red) overall outperforms MSS (green).
States in \minigrid{} encode top-down views of the map with distinct signatures for the agent, walls, and doors, making it easier to learn distance metrics.
In spite of this, the environment remains challenging due to the presence of doors as obstacles and the limited time budget per episode.

\begin{figure}[t]
    \centering
    \includegraphics[width=1\linewidth]{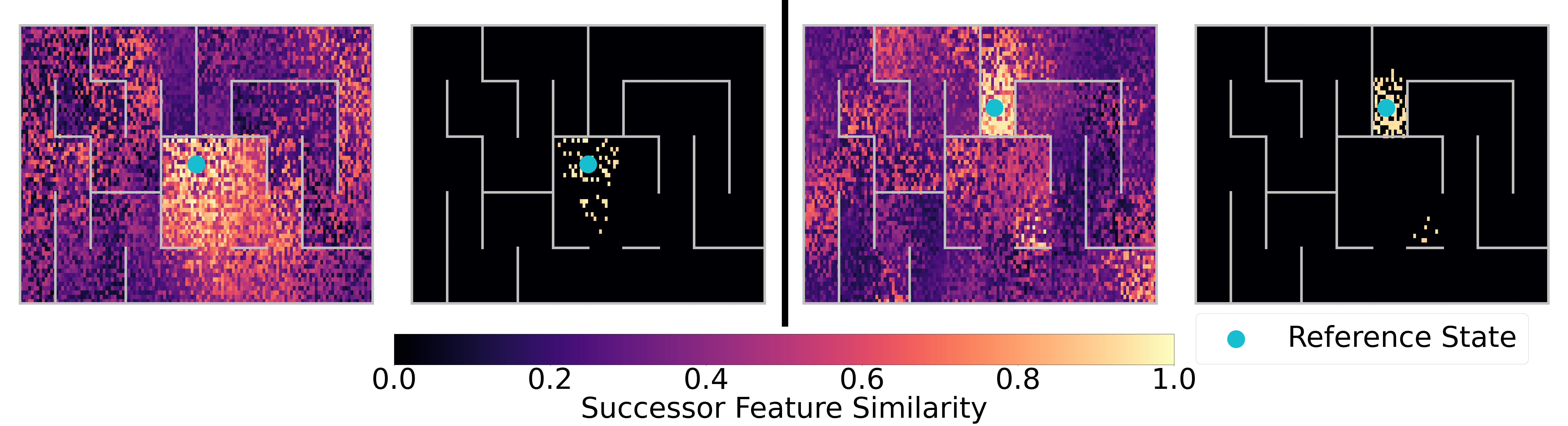}
    \caption{SFS values relative to a reference state (blue dot) in the \textit{Test-1} \vizdoom{} maze. The left two heatmaps use the agent's start state as the reference state while the right two use a distant goal state as the reference state. The first and third (colorful) heatmaps depict all states while the second and fourth (darkened) heatmaps only show states with SFS $> \delta_{\text{local}} = 0.9$.}
    \label{fig:heatmaps}
\end{figure}

\subsection{SFS Visualization}
\label{sec:sfs-viz}
\cutsubsectiondown

\slm primarily relies on SFS and its capacity to approximate geodesic distance imposed by the map's structure.
To provide evidence of this ability, we compute the SFS between a reference state and a set of randomly sampled states.
Figure~\ref{fig:heatmaps} visualizes these SFS heatmaps in a \vizdoom{} maze.
In the first and third panels, we observe that states close to the reference state (blue dot) exhibit higher SFS values while distant states, such as those across a wall, exhibit lower SFS values.
The second and fourth panels show states in which the agent would be localized to the reference state, i.e. states with SFS $> \delta_{\text{local}}$.
With this SFS threshold, we reduce localization errors, thereby improving the quality of the landmark graph.
We provide additional analysis of SFS-derived components, the landmark graph and goal-conditioned policy, in~\Cref{sec:appendix-SLM-analysis}.

\cutsectionup
\section{Conclusion}
\label{sec:conclusion}
\cutsectiondown

In this paper, we presented Successor Feature Landmarks, a graph-based planning framework that leverages a SF similarity metric, as an approach to exploration and long-horizon goal-conditioned RL.
Our experiments in ViZDoom and MiniGrid, demonstrated that this method outperforms current graph-based approaches on long-horizon goal-reaching tasks.
Additionally, we showed that our framework can be used for exploration, enabling discovery and reaching of goals far away from the agent's starting position.
Our work empirically showed that SF can be used to make robust decisions about environment dynamics, and we hope that future work will continue along this line by formulating new uses of this representation.
Our framework is built upon the representation power of SF, which depends on a good feature embedding to be learned.
We foresee that our method can be extended by augmenting with an algorithm for learning robust feature embeddings to facilitate SF learning.

\FloatBarrier

\section*{Acknowledgements}
\label{sec:acknowledgement}
This work was supported by the NSF CAREER IIS 1453651 Grant. JC was partly supported by Korea Foundation for Advanced Studies. WC was supported by an NSF Fellowship under Grant No.\,DGE1418060. Any opinions, findings, and conclusions or recommendations expressed in this material are those of the author(s) and do not necessarily reflect the views of the funding agencies.
We thank Yunseok Jang and Anthony Liu for their valuable feedback.
We also thank Scott Emmons and Ajay Jain for sharing and helping with the code for the SGM \cite{Laskin:2020:SGM} baseline.

\FloatBarrier
\clearpage
\bibliographystyle{plain}
\bibliography{references.bib}

\clearpage
\appendix
\begin{center}
{\huge \textbf{Supplementary Material } }\\
\vspace{4pt}
{\LARGE Successor Feature Landmarks for Long-Horizon Goal-Conditioned Reinforcement Learning}
\end{center}

\label{sec:appendix}
\section{Additional Results}
\label{sec:appendix-results}

\begin{figure}[h]
    \centering
    \bigskip
    \bigskip
    \begin{subfigure}[t]{0.48\linewidth}
        \centering
        \includegraphics[width=\linewidth]{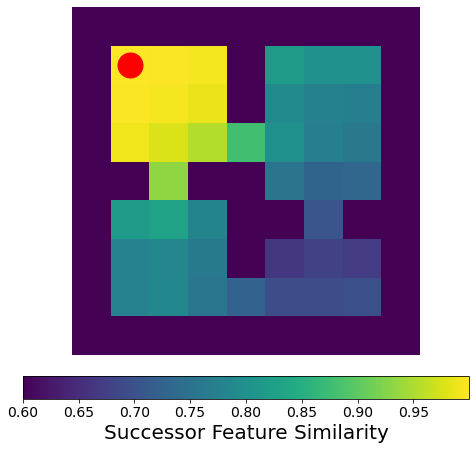}
        \caption{\fourroom{}}
        \label{fig:fourroom-sfs}
    \end{subfigure}
    \hfill
    \begin{subfigure}[t]{0.5\linewidth}
        \centering
        \includegraphics[width=\linewidth]{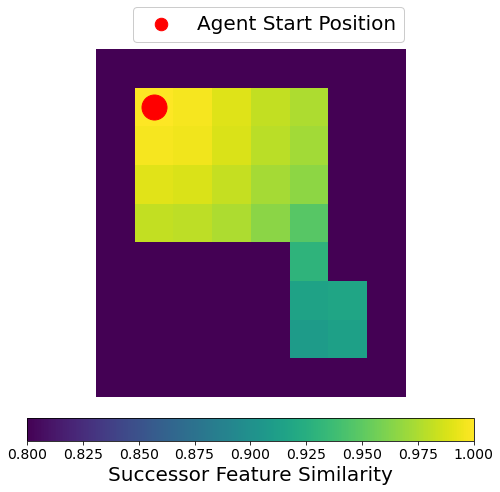}
        \caption{Two-room \multiroom{}}
        \label{fig:two-multiroom-sfs}
    \end{subfigure}
    \hfill
    \vspace{16pt}
    \begin{subfigure}[t]{0.48\linewidth}
        \centering
        \includegraphics[width=\linewidth]{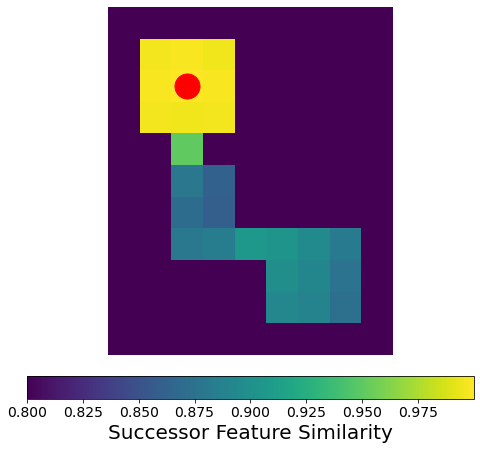}
        \caption{Three-room \multiroom{}}
        \label{fig:three-multiroom-sfs}
    \end{subfigure}
    \hfill
    \begin{subfigure}[t]{0.48\linewidth}
        \centering
        \includegraphics[width=\linewidth]{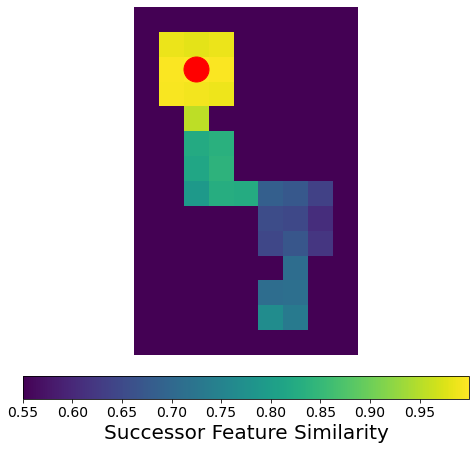}
        \caption{Four-room \multiroom{}}
        \label{fig:four-multiroom-sfs}
    \end{subfigure}
    \vspace{2pt}
    \caption{SFS values relative to the agent's starting state (red dot) for the different \minigrid{} environments.}
    \label{fig:minigrid-sfs}
    \medskip
\end{figure}

\subsection{MiniGrid}
\label{sec:appendix-minigrid-results}
We show visualizations of Successor Feature Similarity (SFS) in the \minigrid{} environment to further illustrate the metric's capacity to capture distance.
Specifically, we compute the SFS between the agent's starting state and the set of possible states and present these values as SFS heatmaps in \Cref{fig:minigrid-sfs} below.
The SFS is distinctly higher for states that reside in the same room as the reference state (red dot). 
Additionally, the SFS values gradually decrease as you move further away from the reference state.
This effect is most clearly demonstrated in the SFS heatmap of Two-room \multiroom{} (top right).

\subsection{ViZDoom}
\label{sec:appendix-vizdoom-results}

We report the standard error for the \textit{random spawn} experiments on \vizdoom{}.
The experiments are run over 5 random seeds.
Note that we use the reported results from the original SGM paper~\cite{Laskin:2020:SGM} for the \textit{SGM-Map} and therefore do not report standard errors.

\bgroup
    \def\arraystretch{1.1}
    \setlength{\tabcolsep}{2pt}
    \vspace{12pt}%
    \begin{table}[h]
        \begin{adjustwidth}{-1in}{-1in}
        \footnotesize
        \begin{center}
            \begin{tabular}{ c | c c c | c c c | c c c } 
             \toprule
          Method & \multicolumn{3}{c|}{\textit{SGM-Map}} & \multicolumn{3}{c|}{\textit{Test-2}} & \multicolumn{3}{c}{\textit{Test-6}}\\
                & Easy & Medium & Hard & Easy & Medium & Hard & Easy & Medium & Hard \\ 
             \midrule %
             Random Actions & $58$\% & $22$\% & $12$\% & $70 \pm 1.2$\% & $39 \pm 1.0$\% & $16 \pm 1.0$ \% & $80 \pm 0.4$\% & $31 \pm 1.0$\% & $18 \pm 0.7$\% \\ 
             Visual Controller & $75$\% & $35$\% & $19$\% & $83 \pm 0.7$\% & $51 \pm 0.5$\% & $30 \pm 0.7$\% & $89 \pm 0.7$\% & $39 \pm 1.1$\% & $20 \pm 1.0$\% \\ 
             SPTM \cite{Savinov:ICLR2018:SPTM} & $70$\% & $34$\% & $14$\% & $78 \pm 0.0$\% & $48 \pm 0.0$\% & $18 \pm 0.0$\% & $88 \pm 0.0$\% & $40 \pm 0.0$\% & $18 \pm 0.0$\% \\ 
             SGM \cite{Laskin:2020:SGM} & \textbf{92}\% & 64\% & 26\% & \textbf{86} $\pm$ \textbf{0.8\%} & $54 \pm 0.7$\% & $32 \pm 0.7$\% & $83 \pm 0.7$\% & $43 \pm 1.2$\% & $27 \pm 1.5$\% \\ %
             \slm [Ours] & \textbf{92} $\pm$ \textbf{0.8\%} & \textbf{82} $\pm$ \textbf{0.6\%} & \textbf{67} $\pm$ \textbf{1.2\%} & $82 \pm 0.7$\% & \textbf{66} $\pm$ \textbf{0.8\%} & \textbf{48} $\pm$ \textbf{1.5\%} & \textbf{92} $\pm$ \textbf{0.6\%} & \textbf{66} $\pm$ \textbf{0.7\%} & \textbf{60} $\pm$ \textbf{0.5\%} \\ %
             \bottomrule %
            \end{tabular}
        \caption{(\textit{Random spawn}) The success rates and standard errors of compared methods on three \vizdoom{} maps.}
        \label{table:random-vizdoom-std}
        \end{center}
        \vspace{6pt}
        \end{adjustwidth}
    \end{table}
\egroup

\section{Ablation Experiments}
\label{sec:appendix-ablations}

We conduct various ablation experiments to isolate and better demonstrate the impact of individual components of our framework.

\subsection{Distance Metric}
\label{sec:metric-ablation}

\bgroup
    \def\arraystretch{1.1}
    \setlength{\tabcolsep}{2pt}
    \vspace{12pt}
    \begin{table}[h]
        \footnotesize
        \begin{center}
            \begin{tabular}{ c | c c c } 
             \toprule
             Metric & Easy & Medium & Hard \\ 
             \midrule
             SFS [Ours] & $92 \pm 1.7$\% & $82 \pm 0.6$\% & $67 \pm 2.6$\% \\
             SPTM's Reachability Network \cite{Savinov:ICLR2018:SPTM} & $83 \pm 1.2$\% & $57 \pm 2.6$\% & $24 \pm 1.9$\% \\
             \bottomrule %
            \end{tabular}
        \caption{The success rates and standard errors of our method and the reachability network ablation on \vizdoom{} \textit{SGM-Map} in the \textit{random spawn} setting. }
        \label{table:metric-ablation}
        \end{center}
        \vspace{6pt}
    \end{table}
\egroup

We compare our SFS metric against the reachability network proposed in SPTM \cite{Savinov:ICLR2018:SPTM} and reused in SGM \cite{Laskin:2020:SGM}. In \Cref{table:metric-ablation}, we observe that SFS outperforms the reachability network on all difficulty levels, indicating that SFS can more accurately represent transition distance between states than the reachability network. The results are aggregated over 5 random seeds.

\subsection{Exploration Strategy}
\label{sec:exploration-ablation}

We investigate the benefit of our exploration strategy, which samples frontier landmarks based on inverse visitation count to travel to before conducting random exploration.
We compare against an ablation which samples frontier landmarks from the landmark set in a uniformly random manner, which is analogous to how SGM \cite{Laskin:2020:SGM} chooses goals in their cleanup step.
We directly measure the degree of exploration achieved by each strategy by tracking state coverage, which we define as the thresholded state visitation count computed over a discretized grid of agent states and report as a percentage over all potentially reachable states.
We report the mean state coverage percentage and associated standard error achieved by the two exploration strategies on the \textit{Test-1} \vizdoom{} map over 5 random seeds. 
Our exploration strategy achieves $79.4 \pm 0.65\%$ state coverage while the uniform random sampling ablation strategy achieves $72.3 \pm 1.90\%$ state coverage, thus indicating that our strategy empirically attains greater exploration of the state space.

\subsection{Landmark Formation}
\label{sec:appendix-landmark-formation}

We compare our progressive building of the landmark set to a clustering scheme akin to the one presented in Successor Options \cite{Ramesh:IJCAI2019:SuccessorOptions}. 
To illustrate the primary benefit of our approach, the ability to track landmark metadata over time, we conduct an experiment with the clustering scheme as an ablation on Three-room \multiroom{}.
Our progressive landmark scheme achieves a mean success rate of $75.0 \pm 14.6\%$ while clustering achieves a rate of $35.6 \pm 18.8\%$, with results aggregated over 5 random seeds.
We observe that our method more than doubles the success rate attained by the clustering method and attribute this outperformance to the beneficial landmark information that we are able to record and utilize in constructing the landmark graph.

We also note a secondary benefit in which landmarks are chosen.
Our approach aims to minimize the distance between chosen landmarks parameterized by $\delta_{add}$ while clustering selects landmarks which are closer to the center of topologically distinguishable regions.
The former method will add landmarks that are far away from existing landmarks, making them more likely to lie on the edge of the explored state space by nature
This in turn can improve exploration via our frontier strategy.
An experiment on \fourroom{} empirically demonstrates this effect, where the average pairwise geodesic distance between landmarks was $6.72 \pm 0.43$ for our method versus $5.72 \pm 0.40$ for clustering.

\cutsectionup
\section{Implementation Details}
\label{sec:appendix-implementation-details}

\subsection{Environments and Evaluation Settings}
\label{sec:appendix-env}

\begin{figure}[h]
    \vspace{12pt}
    \centering
    \begin{subfigure}[t]{0.31\linewidth}
        \centering
        \includegraphics[width=\linewidth]{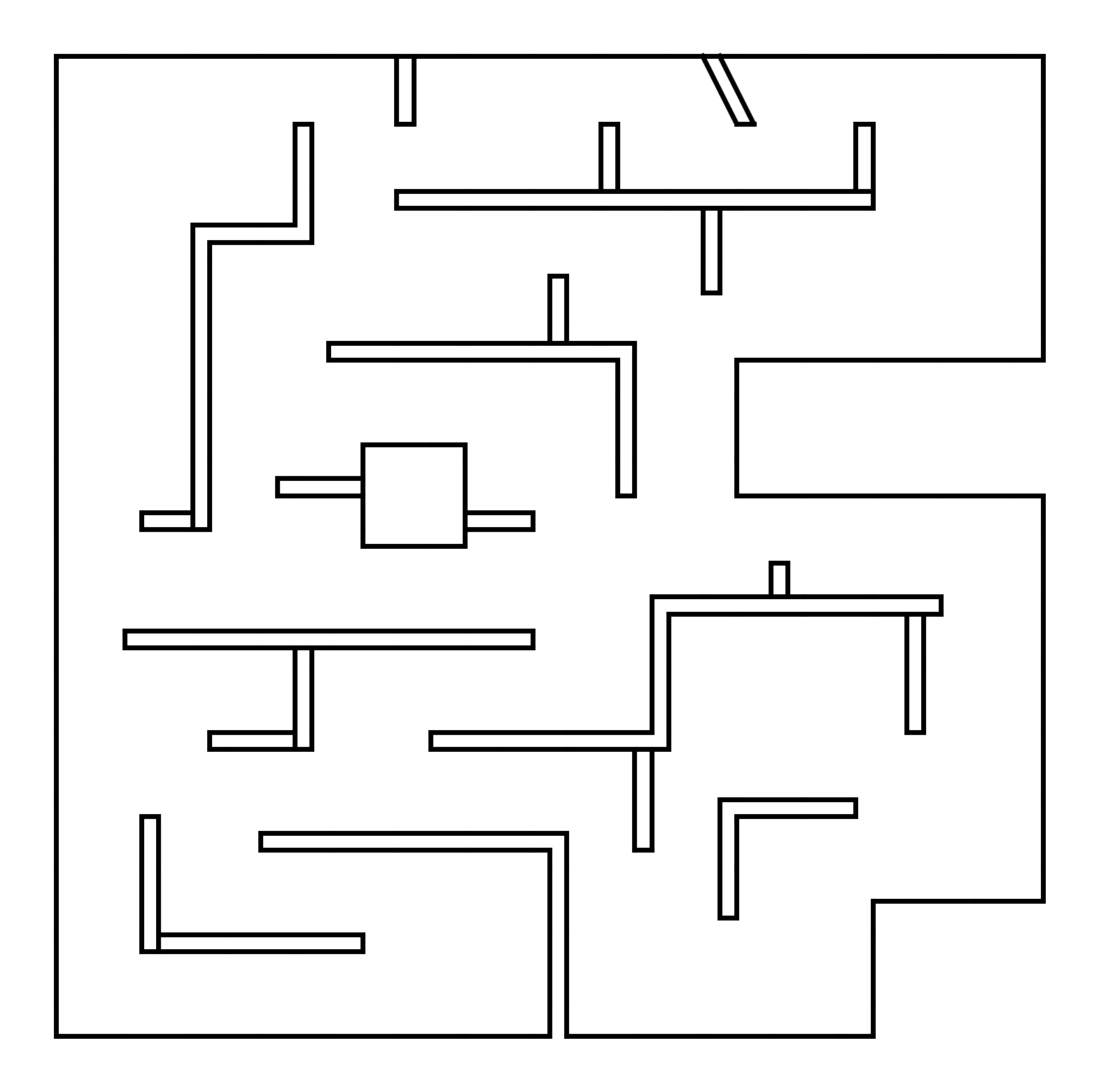}
        \caption{Original SGM map}
        \label{fig:sgm-map}
    \end{subfigure}
    \hfill
    \begin{subfigure}[t]{0.31\linewidth}
        \centering
        \includegraphics[width=\linewidth]{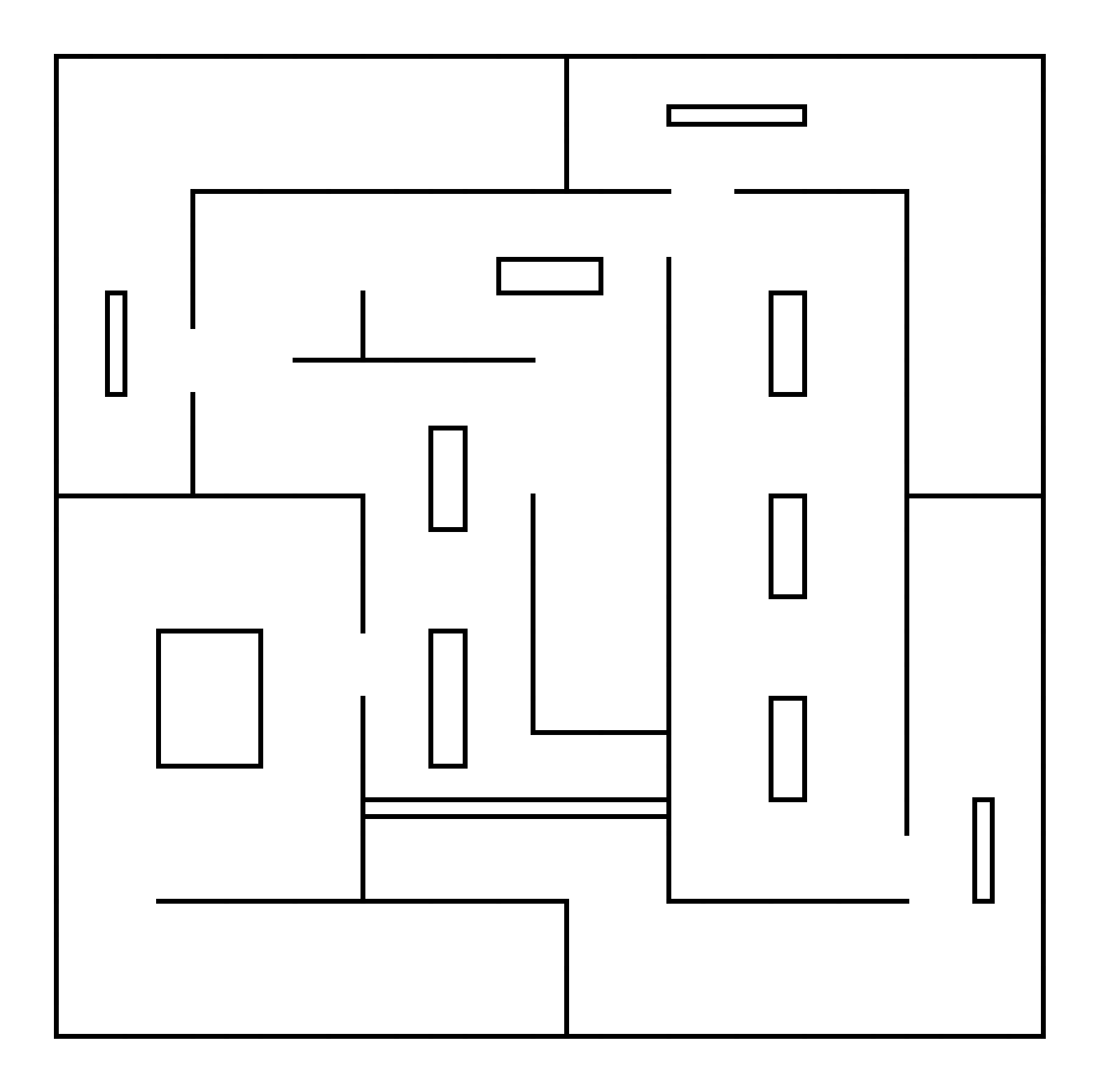}
        \caption{\textit{Test-2}}
        \label{fig:test-2-map}
    \end{subfigure}
    \hfill
    \begin{subfigure}[t]{0.31\linewidth}
        \centering
        \includegraphics[width=\linewidth]{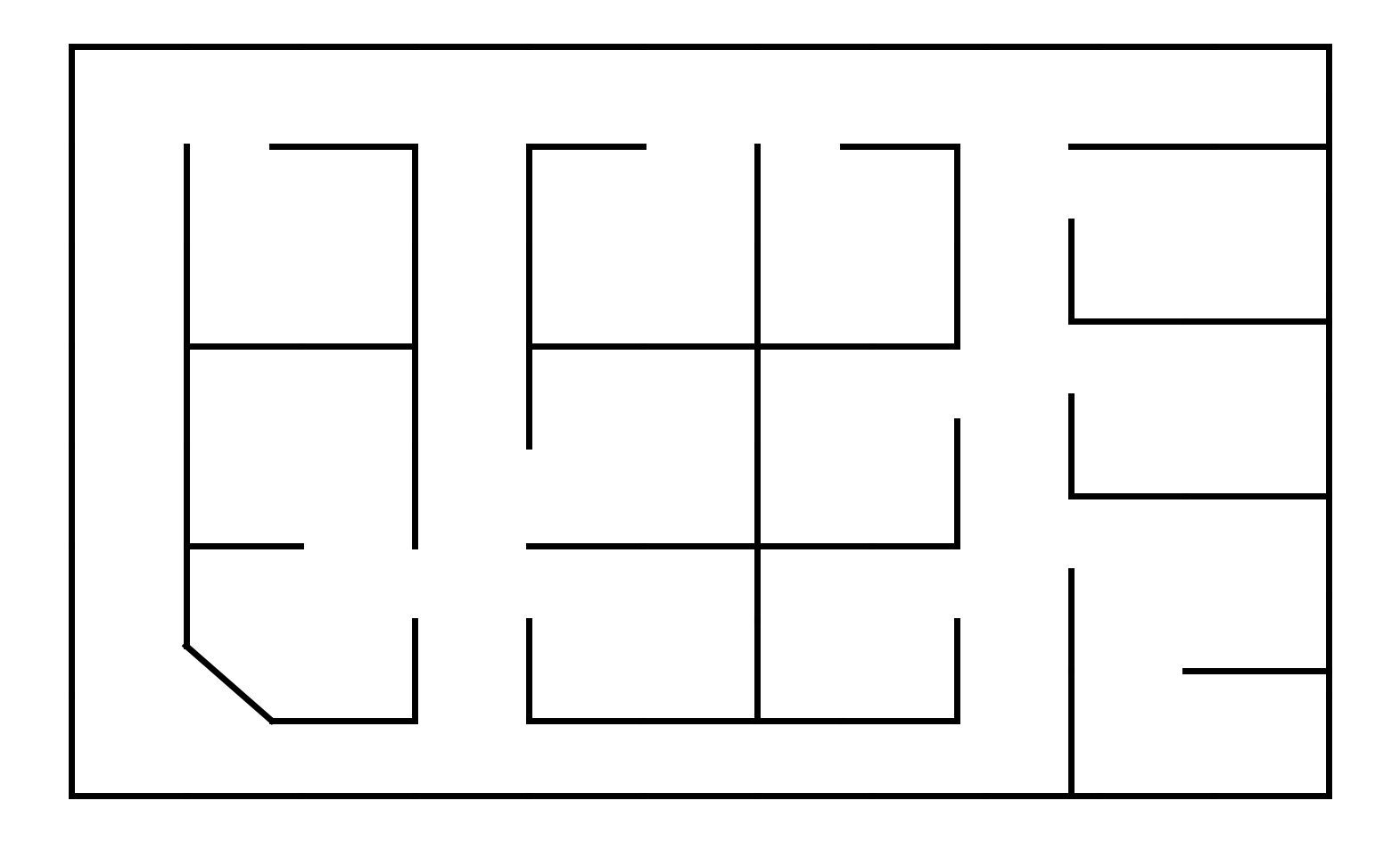}
        \caption{\textit{Test-6}}
        \label{fig:test-6-map}
    \end{subfigure}
    \caption{\vizdoom{} maps used in \textit{random spawn} experiments.}
    \label{fig:random-vizdoom}
    \vspace{12pt}
\end{figure}

\begin{figure}[h]
    \centering
    \begin{subfigure}[t]{0.48\linewidth}
        \centering
        \includegraphics[width=\linewidth]{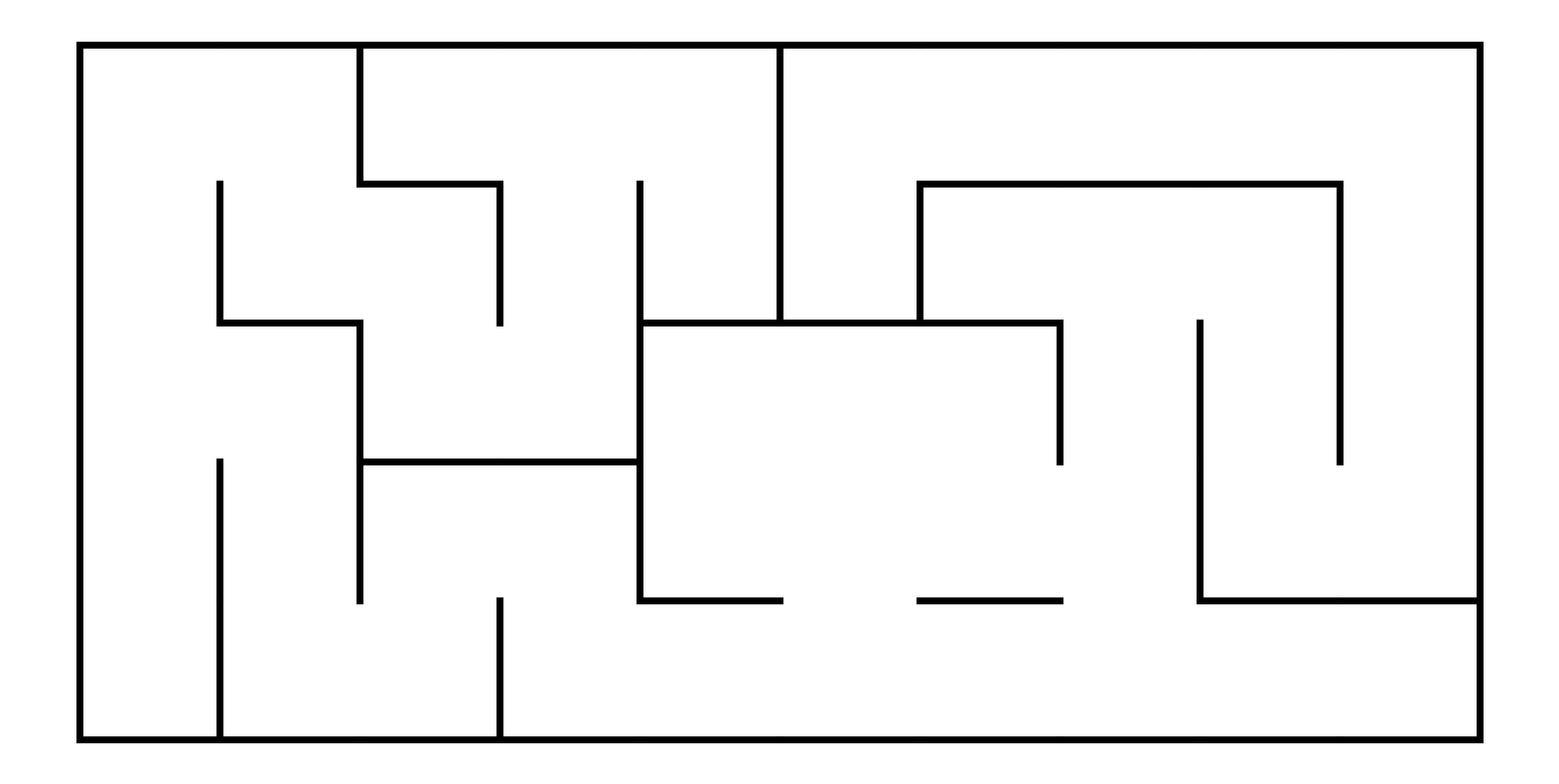}
        \caption{\textit{Test-1}}
        \label{fig:test-1-map}
    \end{subfigure}
    \hfill
    \begin{subfigure}[t]{0.48\linewidth}
        \centering
        \includegraphics[width=\linewidth]{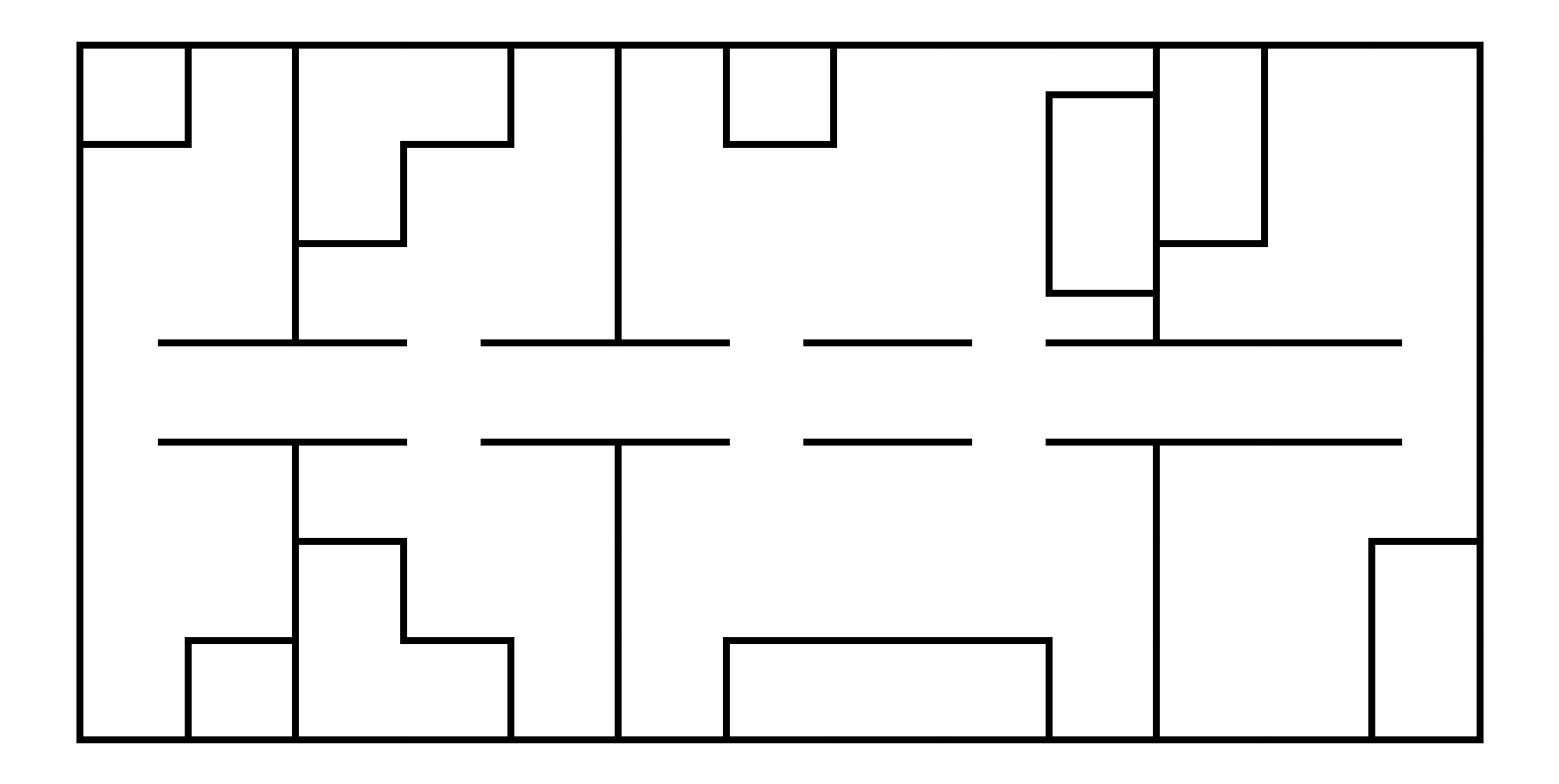}
        \caption{\textit{Test-4}}
        \label{fig:test-4-map}
    \end{subfigure}
    \caption{\vizdoom{} maps used in \textit{fixed spawn} experiments.}
    \label{fig:fixed-vizdoom}
    \vspace{12pt}
\end{figure}

\textbf{ViZDoom:}
The \vizdoom{} visual environment produces $160 \times 120$ RGB first-person view images as observations.
We stacked a history of the 4 most recent observations as our state.
We adopted the same action set as SPTM and SGM: \textit{DO NOTHING, MOVE FORWARD, MOVE BACKWARD, MOVE LEFT, MOVE RIGHT, TURN LEFT, TURN RIGHT}.
As commonly done for \vizdoom{} in previous works, we used an action repetition of 4.
For training, each episode has a time limit of 10,000 steps or 2,500 states after applying action repetition.
We reuse the same texture sets as SPTM and SGM for all mazes.
\Cref{fig:random-vizdoom} shows the maps used in the \textit{random spawn} experiments and \Cref{fig:fixed-vizdoom} shows the maps used in the \textit{fixed spawn} experiments.

During evaluation, the goal state is given as a first-person image observation at the goal position, stacked 4 times.
For both the \textit{random spawn} and \textit{fixed spawn} settings, the success rate for each difficulty level is computed by evaluating the agent on 50 different start-goal pairs generated by each respective sampling procedure.
The reported success rate is the success rate of reaching the goal averaged over 5 random seeds.

\textbf{MiniGrid:}
The \minigrid{} environment provides a compact encoding of a top-down view of the maze as the state.
The encoding represents each object type such as the agent or doors as a distinct 3-length tuple.
The state is $25 \times 25 \times 3$ for \multiroom{} and $9 \times 9 \times 3$ for \fourroom{}.
We use the following action set: \textit{MOVE FORWARD, TURN LEFT, TURN RIGHT, OPEN/CLOSE DOOR}.
For training, each episode has a time limit of 100 steps for \fourroom{} and $n_{\text{rooms}} \cdot 40$ steps for $n_{\text{rooms}}$-room \multiroom{}, where $n_{\text{rooms}}$ is the number of rooms.

During evaluation, the goal state is given as the state if the agent had reached the goal.
The time limits to reach the goal are equivalent to the episode time limits during training.
We compute the success rate of our method and MSS over 100 trajectories each, averaging over 5 random seeds for \fourroom{} and 15 random seeds for \multiroom{}.

\begin{figure}
    \centering
    \begin{subfigure}[t]{0.23\linewidth}
        \centering
        \includegraphics[width=0.9\linewidth]{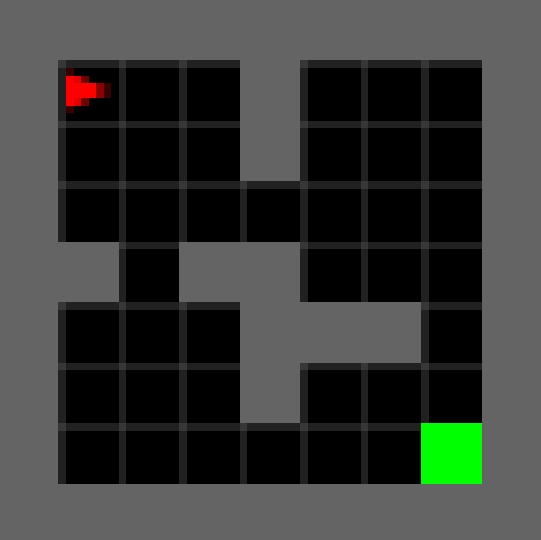}
        \caption{\fourroom{}}
        \label{fig:fourroom}
    \end{subfigure}
    \hfill
    \begin{subfigure}[t]{0.23\linewidth}
        \centering
        \includegraphics[width=0.9\linewidth]{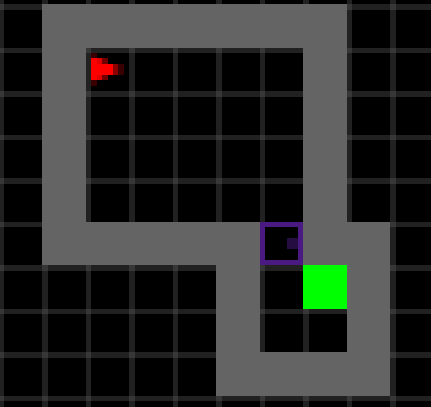}
        \caption{Two-room\\
        \multiroom{}}
        \label{fig:two-multiroom}
    \end{subfigure}
    \hfill
    \begin{subfigure}[t]{0.23\linewidth}
        \centering
        \includegraphics[width=0.9\linewidth]{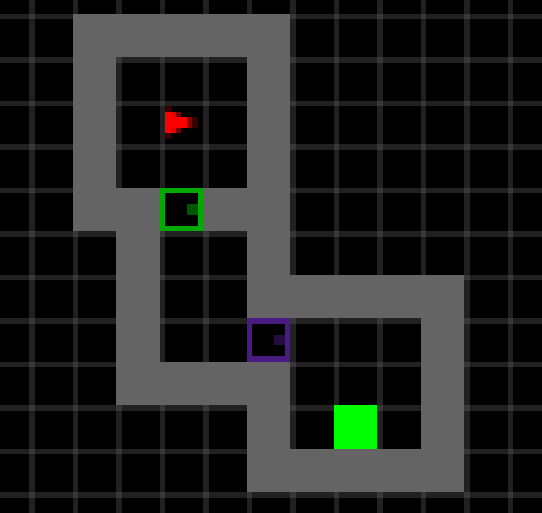}
        \caption{Three-room\\ \multiroom{}}
        \label{fig:three-multiroom}
    \end{subfigure}
    \hfill
    \begin{subfigure}[t]{0.23\linewidth}
        \centering
        \includegraphics[width=0.9\linewidth]{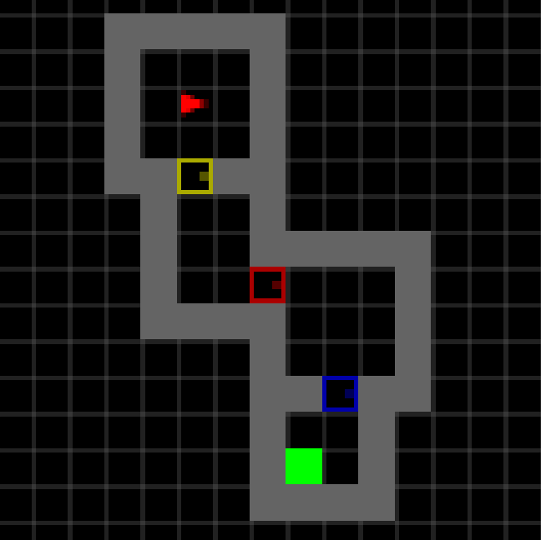}
        \caption{Four-room\\ \multiroom{}}
        \label{fig:four-multiroom}
    \end{subfigure}
    \caption{\minigrid{} maps used in \textit{fixed spawn}. The agent spawns at red arrow and attempts to reach the goal depicted by the green box.}
    \label{fig:minigrid}
\end{figure}

\subsection{Feature Learning}
\label{sec:appendix-feature-learning}
\textbf{ViZDoom:}
For our experiments in \vizdoom{}, we adopt a similar feature learning setup as SGM by reusing the pretrained ResNet-18 backbone from SPTM as a fixed feature encoder.
The network was originally trained with a self-supervised binary classification task: to determine whether a pair of image observations is temporally close within a time step threshold.
The network encodes the image observations as 512-length feature vectors.
Recalling that our state is a stack of the 4 most recent observations, we use the encoder to individually embed each of the observations, and then concatenate the 4 intermediate feature vectors into a 2048-length feature vector.

\textbf{MiniGrid:} For our experiments in \minigrid{}, we learn features by training a convolutional feature encoder with time-contrastive metric learning~\cite{chopra::CVPR2015::metriclearning, Sermanet:ICRA2017:TCN}.
Specifically, we train an encoder $f$ via a triplet loss on 3-tuples consisting of anchor $o^a$, positive $o^p$, and negative observations $o^n$:
\begingroup
    \begin{align}
        \label{eq:time-contrastive}
        ||f(o^a) - f(o^p)||_2^2 + m < ||f(o^a) - f(o^n)||_2^2
    \end{align}
\endgroup

A margin parameter $m = 2$ is used to encourage separation between the (anchor, positive) and the (anchor, negative) pairs.
The 3-tuples are randomly sampled from the replay buffer such that if $o^a$ corresponds to time step $t$, then $o^p$ is uniform randomly sampled from observations from the same episode with time $[t - K_p, t + K_p] = [t - 2, t + 2]$.
Similarly, $o^n$ is randomly sampled from the time intervals, $[t - L_n, t - U_n] \cup [t + U_n, t + L_n] = [t - 15, t - 10] \cup [t + 10, t + 15]$.

The encoder network has the following architecture: two $3 \times 3$ convolutional layers with 32 hidden units, and strides 2 and 1 respectively, each followed by ReLU activations, and ending in a linear layer that outputs 64-dimensional feature vectors.
Additionally, we normalize the feature vector such that $||\phi(o_t)||_2 = \alpha = 10$ following~\cite{Sermanet:ICRA2017:TCN}.
The network is trained using the Adam optimizer \cite{Kingma:ICLR2015:Adam} with a learning rate of $5e-4$ and a batch size of 128.

\subsection{SF Learning}
\label{sec:appendix-sf-learning}
Recall that we estimate SF $\sf$ with a deep neural network parameterized by $\theta: \sf(s, a) \approx \sf_{\theta}(\phi(s), a)$.
Here, $\phi$ is a feature embedding of state.
The parameter $\theta$ is updated via minimizing the temporal difference error~\cite{Kulkarni:2016:DSRL, Barreto:NIPS2017:SuccessorFeatures}:
\begin{align}
    \begin{split}
        &L(s, a, s', \pi, \theta) = %
             \mathbb{E}\left[\Big(\phi(s) + \gamma \widehat{\psi}(\phi(s'), \pi(s')) - \sf_{\theta}(\phi(s), a)\Big)^2\right]
    \end{split}
    \label{eq:td}
\end{align}
where $\widehat{\psi}$ is a target network that is updated at fixed intervals for stability purposes~\cite{Mnih:Nature2015:HumanRL}.
We choose for $\pi$ to be a fixed uniform random policy because we wish for the SF to capture only the structure and dynamics of the environment and not be biased towards any agent behavior induced by a particular policy.
Consequently, we only use transitions from the random policy $\randpi$ for training $\randsf$.

We approximate the SF using a fully-connected neural network, which takes in the features (\Cref{sec:appendix-feature-learning}) as input.
Each hidden layer of the network is followed by a batch normalization and ReLU activation layer.
For updating the parameters of the SF network, we use Adam to optimize on the TD error loss function shown in Eq.~(\ref{eq:td}) with bootstrapped $n-$step transition tuples.
These experience tuples are sampled from a replay buffer of size 100K, which stores trajectories generated by 8 samplers simultaneously running \slm.
For stability, we use a target network $\widehat{\psi}$ that is updated at slower intervals and perform gradient clipping as described in \cite{Mnih:Nature2015:HumanRL}.
\Cref{table:sf-hyperparameters} describes the hyperparameters used for SF learning in each environment.
\setlength{\tabcolsep}{0.25em}
\begin{table}[h]
    \vspace{12pt}
    \footnotesize
    \begin{center}
        \begin{tabular}{ c c c} 
         \toprule
                        & \vizdoom{} & \minigrid{} \\ 
        \textbf{Hyperparameter} & & \\
         \midrule %
         Hidden layer units & 2048, 1024 & 512 \\
         Learning rate & $1e-4$ & $5e-4$ \\
         Batch size & 128 & 128 \\
         $n-$step & 5 & 1 \\
         Replay buffer size & 100K & 20K \\
         Discount $\gamma$ & 0.95 & 0.99 \\
         $\widehat{\psi}$ update interval & 1000 & 250 \\
         Gradient clip $\delta$  & 5 & 1 \\
         \bottomrule %
        \end{tabular}
    \caption{Hyperparameters used in \slm for learning SF for each environment.}
    \label{table:sf-hyperparameters}
    \end{center}
    \vspace{-0.10in}
\end{table}

\begin{table}[h]
    \footnotesize
    \begin{center}
        \begin{tabular}{ c c c} 
         \toprule
                        & \vizdoom{} & \minigrid{} \\ 
        \textbf{Hyperparameter} & & \\
         \midrule %
         $\delta_{add}$ & 0.8 & 0.99\\
         $\delta_{local}$ & 0.9 & 1\\
         $\delta_{edge}$ & - & 1\\
         $N_{front}$ & 1000 & 40 \\
         $N_{explore}$ & 500 & 40 \\
         $\epsilon_{train}$ & 0.05 & 0.1 \\
         $\epsilon_{eval}$ & 0.05 & 0.05 \\
         \bottomrule %
        \end{tabular}
    \caption{Hyperparameters used in \slm's landmark graph, planner, and navigation policy for each environment. See \Cref{sec:appendix-edge-threshold} for details of $\delta_{edge}$ in \vizdoom{}.}
    \label{table:additional-hyperparameters}
    \end{center}
\end{table}

Because the cardinality of \minigrid{}'s state space is much smaller, we restrict \slm to have at most 10 landmarks for \fourroom{} and 30 landmarks for \multiroom{}, which is consistent with the number of landmarks used in MSS as described in \Cref{table:mss-hyperparameters}.

\subsection{SFL Hyperparameters}
\label{sec:appendix-sfl-implementation}
We mainly tuned these hyperparameters: \textit{learning rate}, $\delta_{add}$, and $\delta_{local}$.
In \vizdoom{} for example, we performed grid search over the values of $[10^{-3}, 10^{-4}, 10^{-5}]$ for \textit{learning rate}, $[0.70, 0.75, 0.80, 0.85, 0.90]$ for $\delta_{add}$, and $[0.70, 0.80, 0.9, 0.950]$ for $\delta_{local}$ on the \textit{Train} map.

The best performing values were then used for all other \vizdoom{} experiments. 
We found that our method performed well under a range of values for $\delta_{add}$, and $\delta_{local}$.
In \Cref{table:hyperparameter-tuning}, we report the success rates achieved on \textit{Hard} tasks from a seed-controlled experiment on the \textit{Train} map for random spawn.

\bgroup
    \def\arraystretch{1.1}
    \setlength{\tabcolsep}{2pt}
    \vspace{12pt}
    \begin{table}[h]
        \footnotesize
        \begin{center}
            \begin{tabular}{ c | c c c } 
             \toprule
             $\delta_{add}$ & Success Rate \\
             \midrule
             0.70 & $41\%$ \\
             0.80 & $46\%$ \\
             0.90 & $47\%$ \\
             \midrule
             $\delta_{local}$ & \\ 
             \midrule
             0.80 & $24\%$ \\
             0.90 & $46\%$ \\
             0.95 & $44\%$ \\  
             \midrule
             SGM \cite{Laskin:2020:SGM} & $26\%$ \\
             \bottomrule %
            \end{tabular}
        \caption{The success rates on \textit{Hard} tasks in \textit{Train} \vizdoom{} map for random spawn for varying values of $\delta_{add}$ and $\delta_{local}$. For reference, SGM is included as the best-performing baseline.}
        \label{table:hyperparameter-tuning}
        \end{center}
        \vspace{6pt}
    \end{table}
\egroup

The other hyperparameters were either chosen from related work or not searched over.

\subsection{Optimizations}
\label{sec:appendix-sfl-optimizations}
We perform optimizations on certain parts of \slm for computational efficiency.
To add landmarks, we first store states which pass the SFS add threshold, SFS $< \delta_{add}$, to a candidate buffer.
Then, $N_{cand}$ landmarks are added from the buffer every $N_{add}$ steps.
Additionally, we update the SF representation of the landmarks every $N_{update}$ steps and form edges in the landmark graph every $N_{form-edges}$ steps. 
Finally, we restrict the step-limit for reaching a frontier landmark to be $n_{land}$ times the number of landmarks on the initially generated path so that we do not overly allocate steps for reaching nearby landmarks.

In \vizdoom{}, $N_{cand} = 50$, $N_{add} = 20K$, $N_{update} = 10K$, $N_{form-edges} = 20K$, $n_{land} = 30$.\\
In \minigrid{}, $N_{cand} = 1$, $N_{add} = 3K$, $N_{update} = 1K$, $N_{form-edges} = 1K$, $n_{land} = 8$.

\subsection{Mapping State Space Implementation}
\label{sec:appendix-mss-implementation}
We slightly modify the Mapping State Space (MSS) method to work for our environments.

\textbf{ViZDoom:} Similar to SGM's and our setup, we reuse the pretrained ResNet-18 network from SPTM as a fixed feature encoder $f$. 
The UVFA embeds the start and goal states as feature vectors with this encoder, concatenates them into a 4096-length feature vector, and it them through two hidden layers of size 2048 and 1024, each followed by batch normalization and ReLU activation.
The outputs are the Q-values of each possible action. 
The UVFA is trained with HER, which requires a goal-reaching reward function.
Because states in \vizdoom{} do not directly give the agent's position, we define the reward function based on the cosine similarity between feature vectors given with $f$:

\begingroup
    \setlength\abovedisplayskip{10pt}
    \setlength\belowdisplayskip{10pt}
    \begin{align}
        \label{eq:HER-vizdoom}
        r_t = \mathcal{R}(s_t, a_t, g) = \begin{cases}
                                         0 & f(s_t^{'}) \cdot f(g) > \delta_{reach}  \\
                                         -1 & \text{otherwise}
                                         \end{cases}
    \end{align}
\endgroup

In the function above, we normalize the feature vectors such that $||f(\cdot)||_2 = 1$ and set $\delta_{reach} = 0.8$.

Landmarks are chosen according to farthest point sampling performed in the feature space imposed by the encoder $f$.
During training, the planner randomly chooses a landmark as a goal and attempts to navigate to that goal for 500 steps.
The agent then uses an epsilon greedy policy with respect to the Q-values given by the UVFA for a randomly sampled state as the goal for 500 steps.
It cycles between these two phases until the episode is over.

\textbf{MiniGrid:} The UVFA in \minigrid{} directly maps observations to action q-values.
The UVFA is composed of an encoder with the same architecture as in \Cref{sec:appendix-feature-learning} and a fully-connected network with one hidden layer of size 512 followed by a ReLU activation.
Like in \vizdoom{}, the UVFA encodes the start and goal states, concatenates the feature vectors together, and passes the output through the fully-connected network.
For HER, we reuse the MSS reward function, setting $\delta_{reach} = 1$, i.e. a reward is given only when the agent's next state is the actual goal state.
During training, at every time step, the agent will use the planner with epsilon $\epsilon_{plan}$.
Otherwise, it will use the epsilon greedy policy like in \vizdoom{} above.

\Cref{table:mss-hyperparameters} describes the hyperparameters we used for each environment, which were determined after rounds of hyperparameter tuning.
We give extra attention to the \textit{clip threshold} and \textit{max landmarks} parameters, which MSS~\cite{Huang:NIPS2019:Mapping} mentions are the main hyperparameters of their method.

\begin{table}[t]
    \footnotesize
    \begin{center}
        \begin{tabular}{c @{\hspace{1.5\tabcolsep}} c c c} 
         \toprule
                        & \vizdoom{} & \fourroom{} & \multiroom{} \\ 
         \textbf{Hyperparameter} &  &  & \\ 
         \midrule %
         Learning rate & $10^{-4}$ & $10^{-4}$ & $10^{-3}$ \\
         Batch size & 128 & 256 & 128 \\
         Replay buffer size & 100K & 100K & 100K \\
         Discount $\gamma$ & 0.95 & 0.99 & 0.99 \\
         Target update interval & 100 & 50 & 50 \\
         Clip threshold & -25 & -3 & -5 \\
         Max landmarks & 250 & 10 & 30\\
         HER future step limit & 400 & 10 & 20 \\
         Planner $\epsilon_{plan}$ & - & 0.75 & 0.75 \\
         Explore $\epsilon_{explore}$ & 0.25 & 0.10 & 0.10\\
         Evaluation $\epsilon_{eval}$ & 0.05 & 0.10 & 0.05 \\
         \bottomrule %
        \end{tabular}
    \caption{MSS hyperparameters used for each environment.}
    \label{table:mss-hyperparameters}
    \end{center}
    \vspace{-0.15in}
\end{table}

\subsection{EC-augmented SPTM/SGM Implementation}
\label{sec:appendix-ec-implementation}

We use Episodic Curiosity (EC)~\cite{Savinov:ICLR2019:EC} as an exploration mechanism to enable SPTM and SGM to work in the \textit{fixed spawn} setting on \vizdoom{}.
Specifically, we leverage the exploration abilities of EC to generate trajectories that provide greater coverage of the state space.
SPTM and SGM then sample from these trajectories to populate their memory graphs and to train their reachability and low-level controller networks.
Fortunately, the code repository\footnote{https://github.com/google-research/episodic-curiosity} for EC already has a \vizdoom{} implementation, so minimal changes were required to make it compatible with our experimental setting.

The full procedure is as follows.
First, we train EC on the desired evaluation maze and record the trajectories experienced.
Second, we take a frozen checkpoint of the EC module and use it to generate trajectories for populating the memory graphs.
Third, we train the SPTM/SGM networks on EC's training trajectories recorded in the first step.
Last, we run SPTM/SGM as normal with the constructed memory graph and trained networks.

To make the comparison fair with our method and MSS, we load in the weights from the pretrained ResNet-18 reachability network from SPTM as initial weights for EC's reachability network.
In the first step, we train EC for 2M environment steps, which is the same number of steps we allow for our method to train.
Similar to the \textit{random spawn} experimental setup, we also use the pretrained SPTM/SGM reachability and low-level controller networks, and \textit{fine-tune} them in the third step with 4M steps of training data from the EC-generated trajectories.

\begin{table}[t]
    \footnotesize
    \begin{center}
        \begin{tabular}{c c} 
         \toprule
         \textbf{Hyperparameter} &  \textbf{Value}\\ 
         \midrule %
         Learning rate & $2.5 \times 10^{-4}$ \\
         PPO entropy coefficient & 0.01 \\
         Curiosity bonus scale $\alpha$ &  0.030 \\
         EC memory size & 400 \\
         EC reward shift $\beta$ & 0.5 \\
         EC novelty threshold $b_{novelty}$ & 0.1 \\
         EC aggregation function $F$ & percentile-90\\
         \bottomrule %
        \end{tabular}
    \caption{EC hyperparameters used to generate exploration trajectories for SPTM and SGM.}
    \label{table:ec-hyperparameters}
    \end{center}
    \vspace{-0.15in}
\end{table}

We run validation experiments with EC on the original SGM map to search over the following hyperparameters: curiosity bonus scale $\alpha$, EC memory size, and EC novelty threshold.
We leverage an oracle exploration bonus based on ground-truth agent coordinates as the validation metric.
The same oracle validation metric is used to determine which frozen checkpoint to use in the second step, generating exploration trajectories to populate the memory graph.
For the other hyperparameters, we reuse the values chosen for \vizdoom{} in the original EC paper~\cite{Savinov:ICLR2019:EC}.
\Cref{table:ec-hyperparameters} describes the hyperparameters that we selected for EC.

\subsection{Computational Resources}
Each \slm run uses a single GPU and we use 4 GPUs in total to run the \slm experiments: 2 NVIDIA GeForce GTX 1080 and and 2 NVIDIA Tesla K40c.
The code was run on a 24-core Intel Xeon CPU @ 2.40 GHz.

\subsection{Asset Licenses}
Here is a list of licenses for the assets we use:
\begin{enumerate}
    \item \textit{rlpyt} code repo: MIT license
    \item \vizdoom{} environment: MIT license
    \item \minigrid{} environment: Apache 2.0 license
    \item MSS code repo: MIT license
    \item SPTM code repo: MIT license
    \item SGM code repo: Apache 2.0 license
    \item EC code repo: Apache 2.0 license
\end{enumerate}

\section{Tackling Perceptual Aliasing in ViZDoom}
\label{sec:appendix-aliasing}
\begin{figure}[h]
    \vspace{12pt}
    \centering
    \includegraphics[width=1\linewidth]{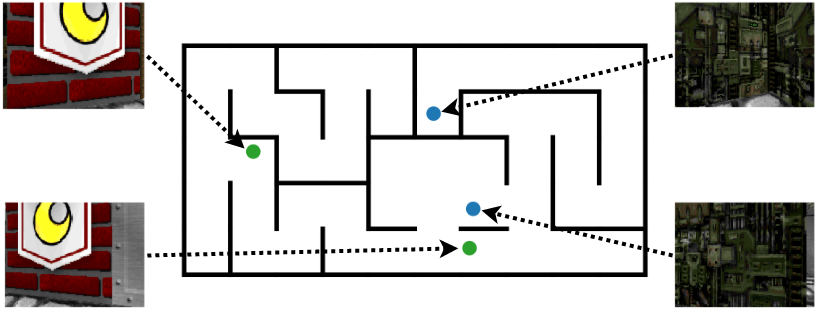}
    \caption{Examples of perceptual aliasing in \vizdoom{}. Same colored dots correspond to pairs of perceptually aliased observations which have pairwise SFS values $> \delta_{local} = 0.9$.}
    \label{fig:perceptual-aliasing}
\end{figure}

\cutsectiondown
Perceptual aliasing is a common problem in visual environments like \vizdoom{}, where two image observations look visually similar, but correspond to distant regions of the environment.
This problem can cause our agent to erroneously localize itself to distant landmarks, which in turn harms the accuracy of the landmark graph and planner.
Figure~\ref{fig:perceptual-aliasing} gives examples of the perceptual aliasing problem where the pairs of visually similar observations also have very high SFS values relative to each other.
We take several steps to make \slm more robust to perceptual aliasing, as described in the following sections.

\subsection{Aggregating SFS over Time}
\label{sec:appendix-sfs-aggregation}
We adopt a similar procedure used in SGM and SPTM \cite{Laskin:2020:SGM, Savinov:ICLR2018:SPTM} to make localization more robust by aggregating SFS over a short temporal window.
Specifically, we maintain a history of SFS values over the past $W$ steps, and output the median as the final SFS value $S$.
This is defined as follows:
\begingroup
    \setlength\abovedisplayskip{12pt}
    \begin{equation}
        S = \text{median} ([\sfs(s_{t - W}, \cdot), ..., \sfs(s_t, \cdot)])
    \end{equation}
\endgroup

where $\sfs(s_t, \cdot)$ is the $|L|$-length vector containing $\sfs(s_t, l), \forall l \in L$. 
The median function is taken over the time dimension such that $S$ is length $|L|$.
Then, for example, if we were attempting to add $s_t$ as a landmark, we would compute $l_t = \argmax_{l\in L} S_l$ and check if $S_{l_t} < \delta_{add}$ to decide if we should add $s_t$ as a landmark. 
For our experiments, we set $W = 8$.

\subsection{Edge Threshold}
\label{sec:appendix-edge-threshold}
We give special consideration to the edge threshold $\delta_{edge}$ in \vizdoom{} due to the larger scale of the environment and recognized that a fixed edge threshold may not work depending on the stage of training.
For example, a low value of $\delta_{edge}$ would allow connections to form in the early stages of training, but would also introduce unwanted noise to the graph as the number of nodes grows.
Therefore, we wish for $\delta_{edge}$ to dynamically change depending on the status of the graph. 
We define it as follows:
\begingroup
    \setlength\abovedisplayskip{12pt}
    \begin{equation}
        \delta_{edge} = \underset{l_{i} \rightarrow l_{j}}{\text{median}}(N_{l_{i}\rightarrow l_{j}}^l)
    \end{equation}
\endgroup

In other words, an edge is formed if the number of landmark transitions on that edge is greater than the median number of landmark transitions from all edges.
We found that the median is a suitable threshold for enabling sufficient graph connectivity in the beginning while also reducing the the number of erroneous edges as the graph grows in scale.

\subsection{Edge Filtering}
\label{sec:appendix-edge-filtering}
We apply edge filtering to the landmark graph to reduce the number of potential erroneous edges.
In \textit{random spawn} experiments, we adopt SGM's $k$-nearest edge filtering where for each vertex, we only keep edges that have the top $k = 5$ number of transitions from that vertex.
In \textit{fixed spawn} experiments, we instead introduce \textit{temporal} filtering where we only keep edges between landmarks that were added during similar time periods.
Specifically, assuming landmarks are labeled by the order in which they are added, $uv \in G \rightarrow |u - v| < \tau_{temporal} \cdot |L|$, where $L$ is the number of landmarks.
Our intuition for \textit{temporal} filtering is based on how the agent in the \textit{fixed spawn} setting will add new landmarks further and further away from the starting point as it explores more of the environment over time.
Because the agent must pass by the most recently added landmarks at the edge of the already explored areas in order to add new landmarks, landmarks added within similar time periods are overall likely to be closer together.
In our experiments, we set $\tau_{temporal} = 0.1$.

Additionally, we adopt the cleanup step proposed in SGM by pruning failure edges. 
If the agent is unable to traverse an edge with the navigation policy, we remove that edge.
To account for the case where the edge is correct, but the goal-conditioned policy has not yet learned how to traverse that edge, we "forget" about failures over time such that only edges that are \textit{repeatedly} untraversable are excluded from the graph.
The agent forgets about failures that occurred over 80K steps ago.

\begin{figure}[h]
    \centering
    \includegraphics[width=1\linewidth]{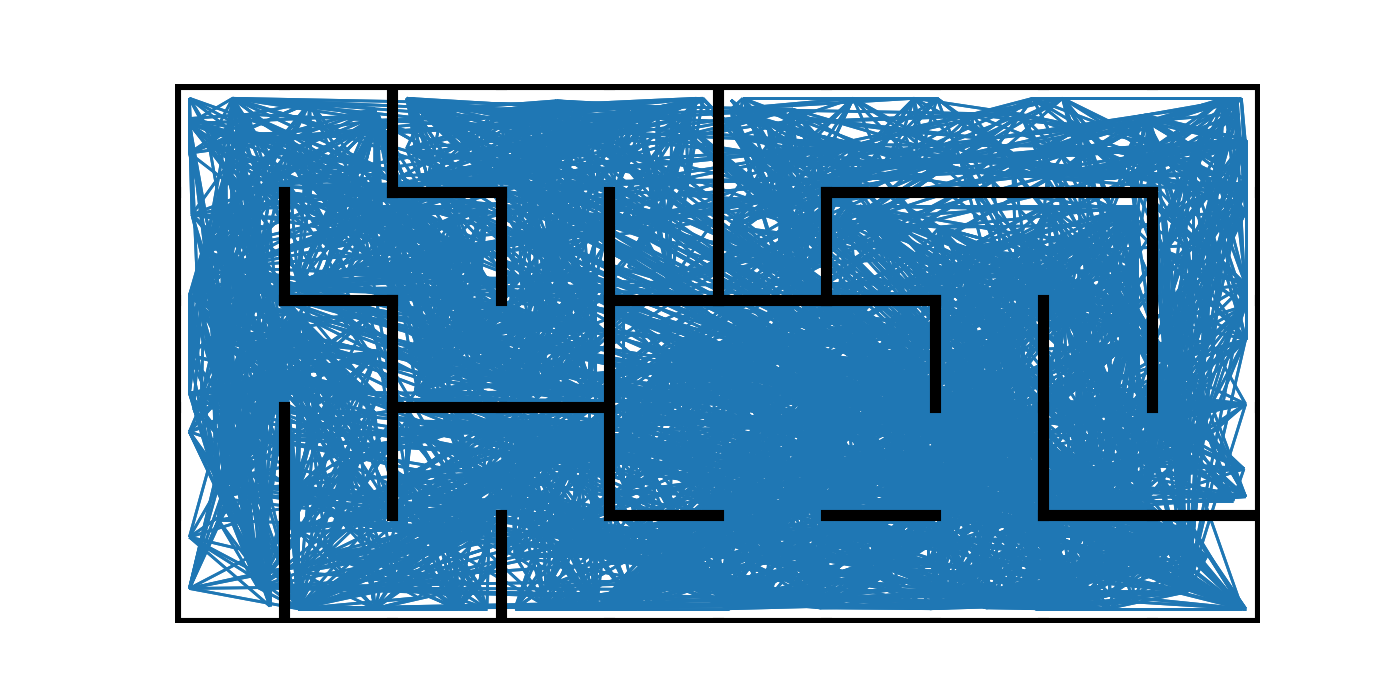}
    \caption{Graph formed with only empirical landmark transitions: $|L| = 517, |E| = 8316$.}
    \label{fig:no-filter}
\end{figure}

\begin{figure}[h]
    \centering
    \includegraphics[width=1\linewidth]{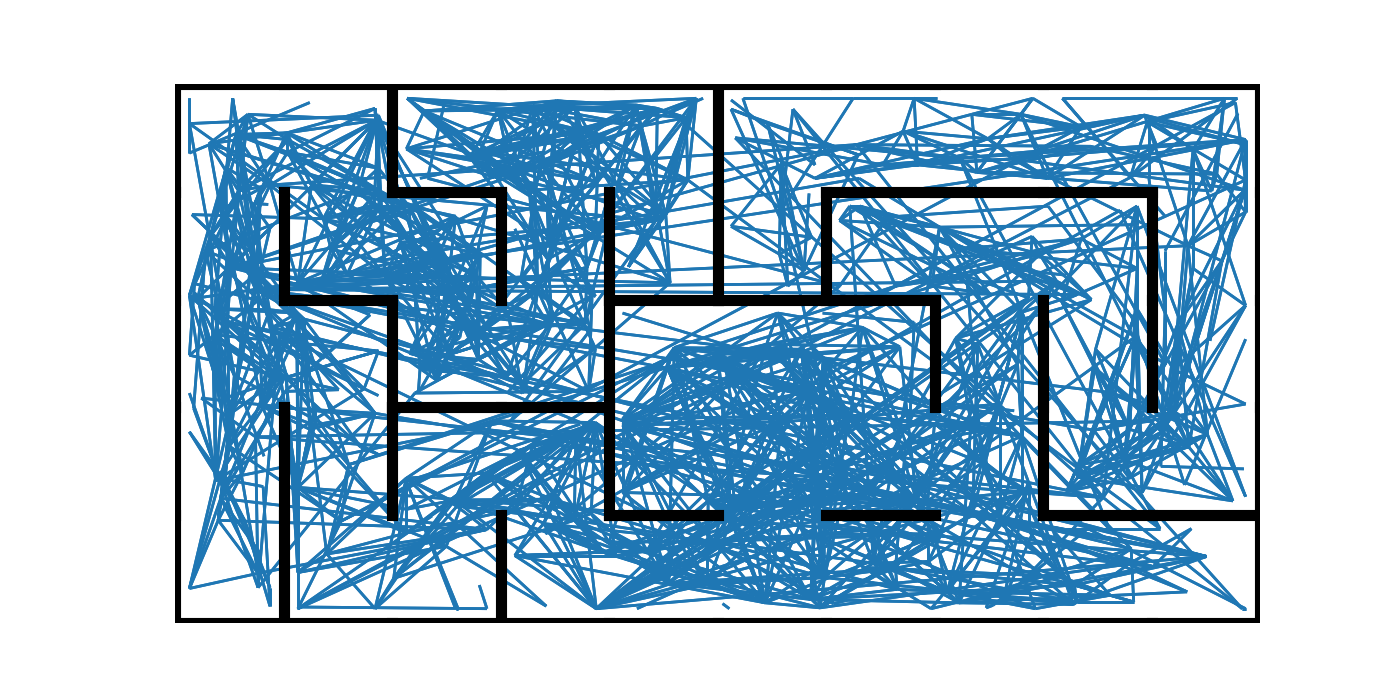}
    \caption{Graph formed with empirical landmark transitions, \textit{temporal filtering}, and failure edge cleanup: $|L| = 517, |E| = 2984$.}
    \label{fig:filter}
\end{figure}

These procedures can improve the quality of the landmark graph.
Figure~\ref{fig:no-filter} shows the landmark graph formed in one of our \vizdoom{} \textit{fixed spawn} experiments when no additional edge filtering steps are used. 
The graph has many incorrect edges which connect distant landmarks.
On the other hand, Figure~\ref{fig:filter} shows the landmark graph with \textit{temporal} filtering and failure edge cleanup.
These procedures eliminate many of the incorrect edges, resulting in a graph which respects the wall structure of the maze to a much higher degree.

\vspace{10pt}

\section{\slm Analysis}
\label{sec:appendix-SLM-analysis}

We conduct further analysis on the components of \slm to study how each contributes to the agent's exploration, goal-reaching, and long-horizon planning abilities.

\subsection{Exploration}
\label{sec:appendix-exploration-analysis}
For our \textit{fixed spawn} experiments, as the agent progresses in training, it should spend more time in faraway areas of the state space.
In Figure~\ref{fig:landmark-visitations}, we show that our agent exhibits this behavior. 
Each landmark is colored by the relative rank of its visitations, where a lighter color corresponds to spending more time at a landmark.
Early in training (top), the agent has discovered some faraway landmarks, but does not spent much time in these distant areas as indicated by the darker color of these landmarks.
Later in training (bottom), the agent has both added more landmarks and spent more time near distant landmarks.
This is also shown by how the lighter colors are more evenly distributed across the map.
We expect agents without effective exploration strategies to remain near the center of the map.

\subsection{Goal-Conditioned Policy}
\label{sec:appendix-policy-analysis}
\begin{figure}[!t]
    \centering
    \hfill
     \begin{subfigure}[b]{0.48\linewidth}
        \centering
        \includegraphics[width=\linewidth]{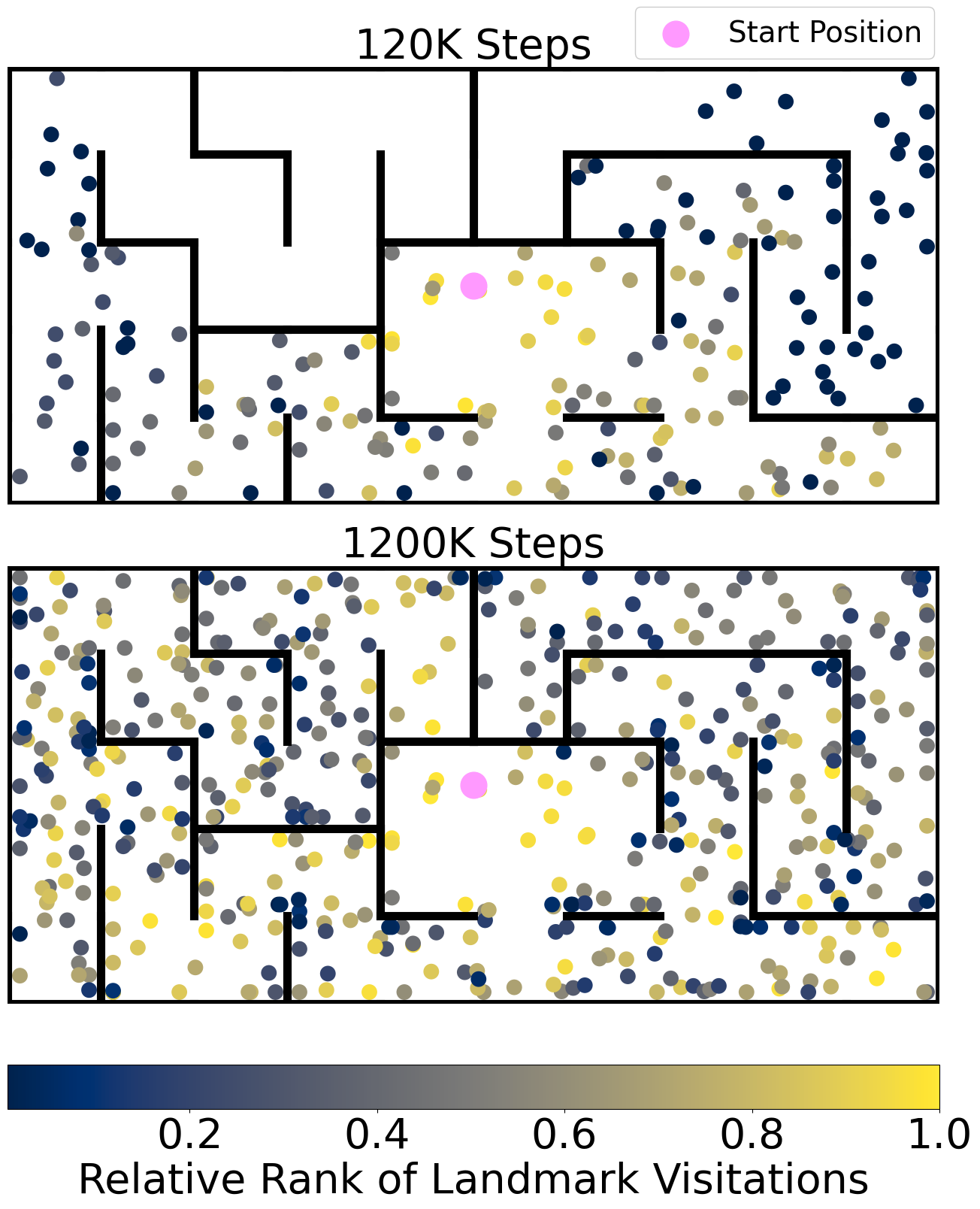}
        \caption{ }
        \label{fig:landmark-visitations}
     \end{subfigure}
    \hfill
     \begin{subfigure}[b]{0.485\linewidth}
        \centering
        \includegraphics[width=\linewidth]{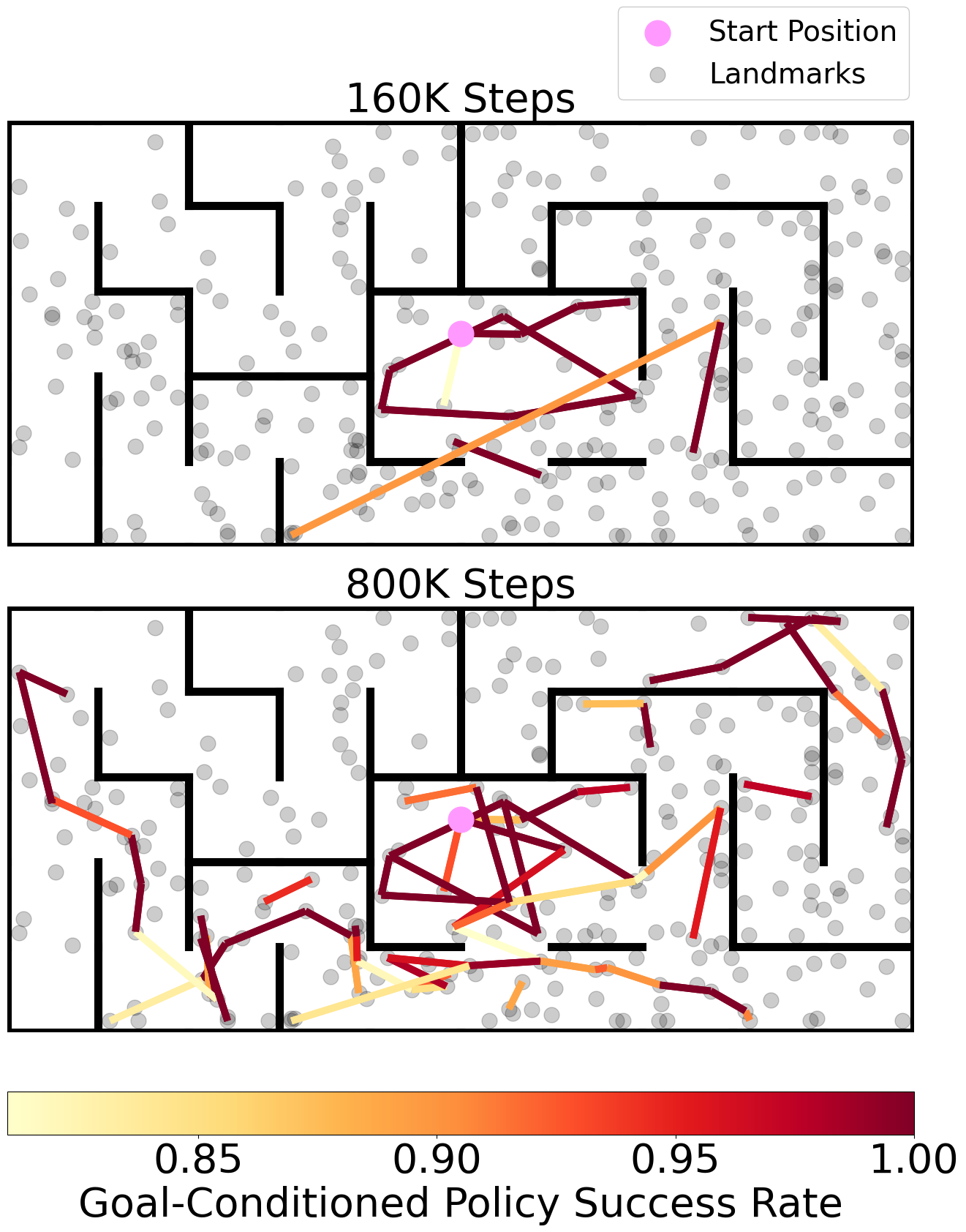}
        \caption{ }
        \label{fig:local-policy}
    \end{subfigure}
        \hfill
    \vspace*{-8pt}
    \caption{Visualizations of exploration (left) and goal-conditioned policy (right) on the \textit{fixed spawn} \textit{Test-1 }\vizdoom{} maze. \textbf{Left:} Each landmark colored according to the relative rank of its visitations at different time steps. \textbf{Right:} Landmark edges colored by goal-conditioned policy success rate at different time steps. Only edges with $\ge 80\%$ success rate are shown.}
    \vspace*{-0.1in}
\end{figure}

We study how the goal-conditioned policy improves over training.
Figure~\ref{fig:local-policy} shows how the goal-conditioned policy's success rate over certain landmark edges increases over time, with success defined as reaching the next landmark within 15 steps.
On top, we see that the policy is only accurate for edges near the start position during an early stage of training.
Additionally, there is an incorrect, extra long edge that is most likely attributed to localization errors that will be corrected later on as the agent further trains its SF representation.
On the bottom, we observe that the goal-conditioned policy improves in more remote areas after the agent has explored more of the maze and completed more training.

\subsection{Planning with Landmark Graph}
\label{sec:appendix-planning-analysis}
\begin{figure}[h]
    \centering
    \includegraphics[width=\linewidth]{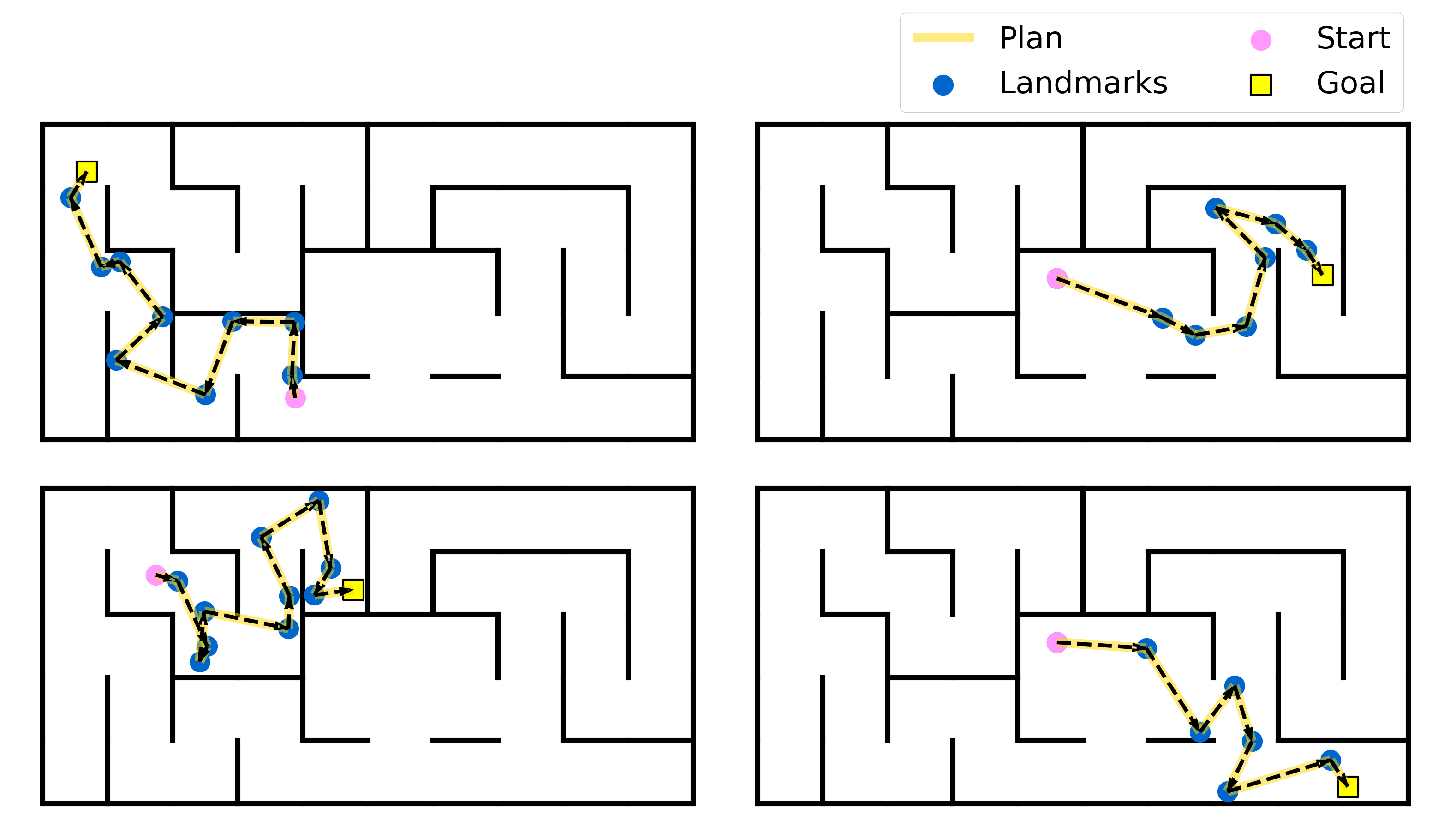}
    \caption{Examples of planned paths for various start (pink dot) and goal (yellow square) locations. The paths are formed by conducting planning on the landmark graph. These examples are taken from one of the \textit{fixed spawn} experiments on the  \textit{Test-1} map.}
    \label{fig:fixed-spawn-paths}
\end{figure}

Here, we look at the long-horizon landmark paths that the \slm agent plans over the graph.
Examples of planned paths are shown in Figure~\ref{fig:fixed-spawn-paths}.
We observe that the plans accurately conform to the maze's wall structure.
Additionally, consecutive landmarks in the plan are not too far apart, which helps the success rate of the goal-conditioned policy because SFS is more accurate when the start-goal states are within a local neighborhood of each other.
We acknowledge that the planned paths can be longer and more ragged than the optimal shortest path to the goal.
The partial observability of the environment is one primary reason for this, where there are multiple first-person viewpoints per (x, y) location. 
For example, two landmarks may be relatively closer in terms of number of transitions even if they appear further away on the top-down map because they both share a $30^{\circ}$ view orientation.

\section{Mapping State Space Analysis}
\label{sec:appendix-MSS-analysis}
In this section, we offer additional analysis on Mapping State Space (MSS) and elaborate on potential reasons why the method struggles to achieve success in our environments.
Our study is conducted in the context of \textit{fixed spawn} experiments in \vizdoom{}.

\subsection{UVFA}
\begin{figure}[!h]
    \centering
    \includegraphics[width=0.9\linewidth]{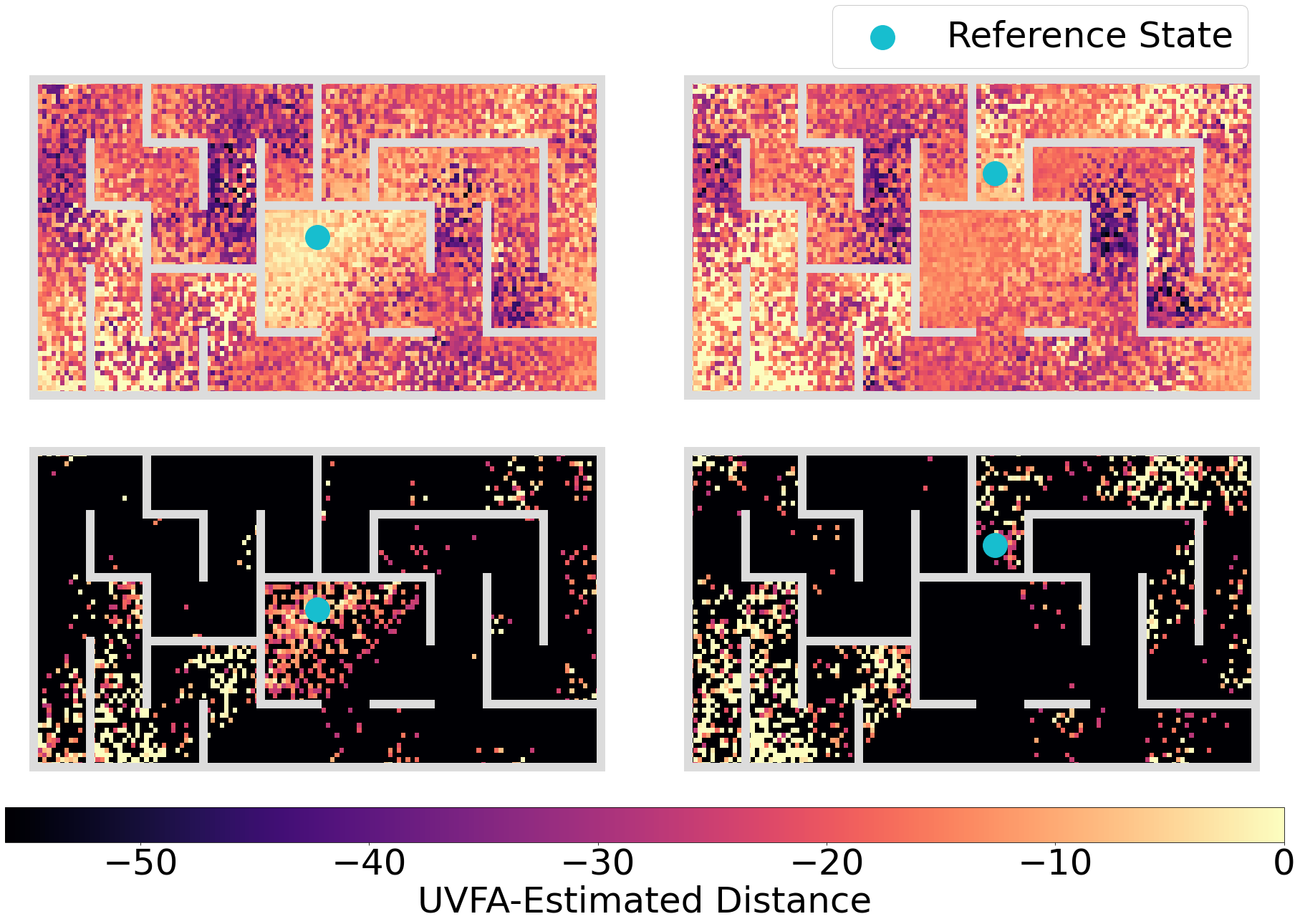}
    \caption{Distances estimated with MSS' UVFA relative to a reference state (blue dot) in the \textit{fixed spawn} \vizdoom{} maze. The left column uses the agent's start state as the reference state while the right column uses a distant goal state as the reference state. The top row depicts all states while the bottom row shows states with distance $>= -5$, the edge clip threshold. States that do not pass the threshold, i.e. distance $< -5$ are darkened.}
    \label{fig:uvfa-no-gt-heatmaps}
\end{figure}

The UVFA is central to the MSS method, acting as both an estimated distance function and a local goal-conditioned navigation policy.
First, we qualitatively evaluate how well the UVFA estimates the distance between states by creating heatmaps similar to those in Figure~\ref{fig:heatmaps}.
We use a trained UVFA to estimate the pairwise distances between a reference state and a random set of sampled states, and plot those distances in Figure~\ref{fig:uvfa-no-gt-heatmaps}.
The top row shows the distance values for all states; the bottom row shows the distance values for states which clear the edge clip threshold $= -5$ and consequently would have edges to the reference state.
We choose a edge clip threshold smaller than the one used in our experiments to reduce the number of false positives and for ease of visualization.
We observe that the estimated distances are noisy overall, where many states far away from the reference state are given small distances which pass the edge clip threshold.
These errors cause the landmark graph to have incorrect edges that connect distant landmarks
As a result, the UVFA-based navigation policy cannot accurately travel between these distant pairs of landmarks.

We hypothesize that the UVFA is inaccurate because the learned feature space does not perfectly capture the agent's (x, y) coordinates for localization.
This causes errors in the HER reward function where it may give a reward in cases when the agent has not yet reached the relabeled goal state.
This is exacerbated by the perceptual aliasing problem.
With this noisy reward function, the UVFA learning process becomes very challenging.

\begin{figure}[!h]
    \centering
    \includegraphics[width=0.9\linewidth]{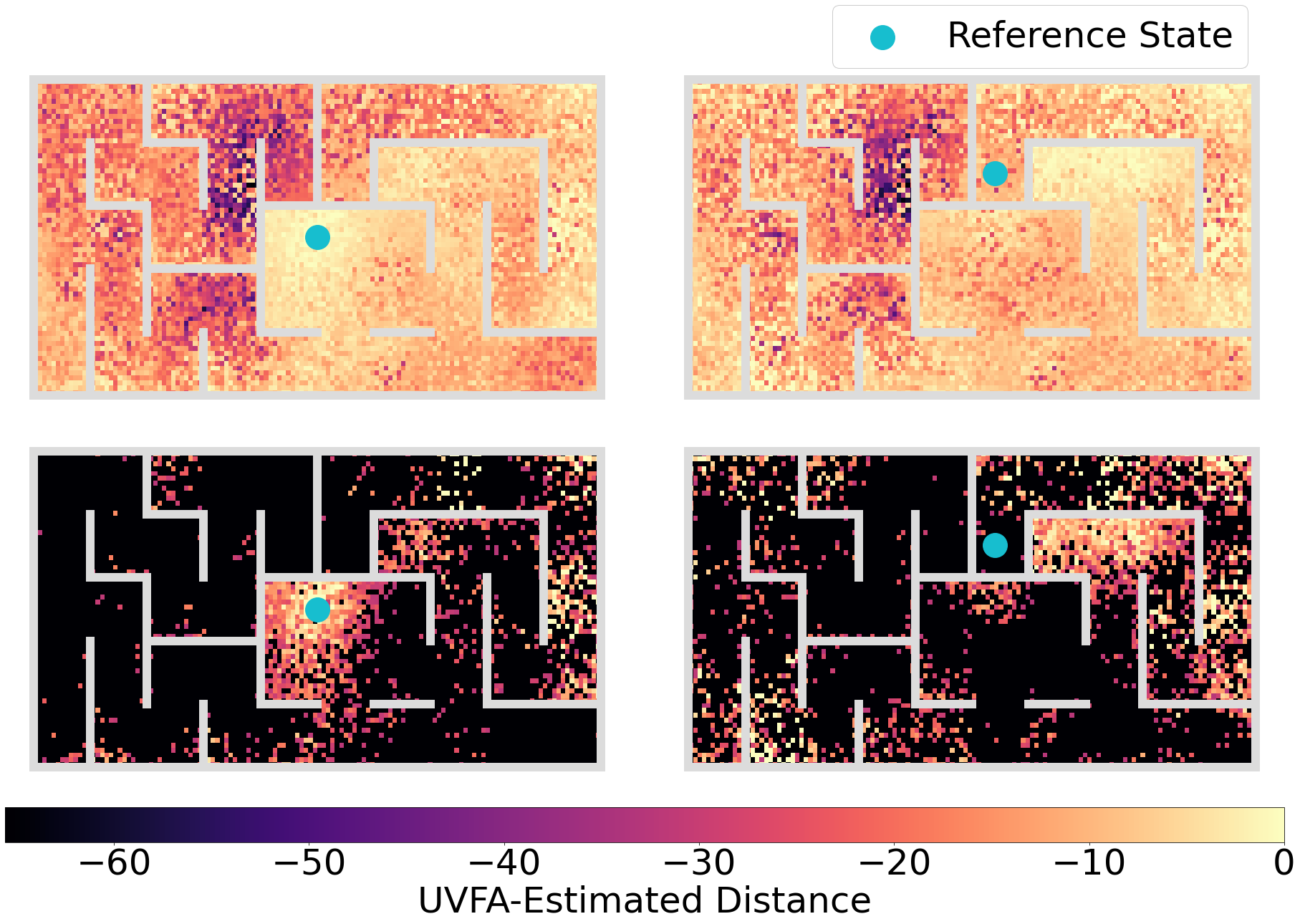}
    \caption{Distances estimated with MSS' UVFA relative to a reference state (blue dot) in the \textit{fixed spawn} \vizdoom{} maze. We assume a similar setup as Figure~\ref{fig:uvfa-no-gt-heatmaps}, but the UVFA is trained using HER with ground-truth (x, y) coordinate data.}
    \label{fig:uvfa-gt-heatmaps}
\end{figure}

For further analysis, we re-train the UVFA with a HER reward function that is given the agent's ground-truth (x, y) coordinates.
Now, the agent is only given a reward when it exactly reaches the relabeled goal state.
We recreate the same UVFA-estimated distance heatmaps using this training setup, shown in Figure~\ref{fig:uvfa-gt-heatmaps}.

We see in the left column that states which pass the edge clip threshold are now more concentrated near the reference state.
However, the estimated distances are still very noisy, especially in the right column where the reference state is in a distant location.
We believe that learning an accurate distance function remains difficult because the features inputted into the UVFA only give a rough estimate of the agent's (x, y) position rather than capture it fully.

\subsection{Landmarks}
\begin{figure}[!t]
    \medskip
    \centering
    \includegraphics[width=0.9\linewidth]{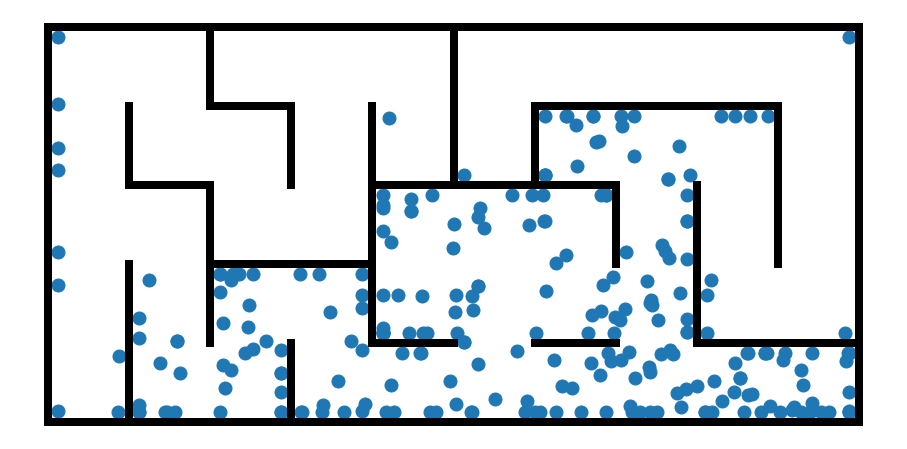}
    \caption{Landmarks selected using FPS in a MSS \textit{fixed spawn} experiment.}
    \label{fig:mss-landmarks}
    \medskip
\end{figure}

MSS uses a set of landmarks to represent the visited state space. 
From a random subset of states in the replay buffer, the landmarks are selected using the farthest point sampling (FPS) algorithm, which is intended to select landmarks existing at the boundary of the explored space. 
Figure~\ref{fig:mss-landmarks} shows the landmarks selected near the end of training in a MSS \textit{fixed spawn} experiment.
We note that when we increase the max number of landmarks in our experiments, the observed performance either decreases or remains the same.

The set of landmarks only partially covers the maze, thereby limiting the agent's ability to reach and further explore distant areas.
Because FPS operates in a learned feature space, the estimated distances between potential landmarks can be noisy, leading to inaccurate decisions on which landmarks should be chosen. 
Furthermore, MSS does not explicitly encourage the agent to navigate to less visited landmarks.
States in unexplored areas become underrepresented in the replay buffer and therefore, are unlikely to be included in the initial random subset of states from which landmarks are chosen.

\section{Episodic Curiosity Analysis}
\label{sec:appendix-ec-analysis}

In this section, we conduct additional experiments regarding the Episodic Curiosity (EC) augmented SPTM and SGM baselines to better understand the benefits and shortcomings of these combined methods.
This study is completed within the context of \textit{fixed spawn} experiments in \vizdoom{}.

We run ablations of the EC + SPTM and EC + SGM baselines on the \textit{Test-1} map within the \textit{fixed spawn} evaluation setting.
Two components of SPTM/SGM are changed in the ablations: the high-level graph, and the reachability and locomotion networks.
Specifically, we vary how we collect the trajectories used to populate the graph and to train the two networks.
The trajectories are generated by the following agent variations:
\begin{enumerate}[labelindent=0cm,labelsep=3pt,noitemsep,nolistsep, topsep=0pt,leftmargin=*]
    \item \textbf{Fixed start (FS)}: randomly acting agent that begins each episode in the \textit{same} starting location.
    \item \textbf{Random start (RS)}: randomly acting agent that begins each episode in a starting location that is sampled uniform randomly across the map.
    \item \textbf{EC}: agent running the EC exploration module and begins each episode in the \textit{same} starting location.
\end{enumerate}

For populating the graph, the agents build and sample from a replay buffer containing trajectories of 100 episodes of 200 steps each, following the setup of SGM~\cite{Laskin:2020:SGM}.
The EC agent uses a frozen checkpoint of its reachability and policy networks when collecting these trajectories.
The setup for training the reachability and locomotion networks of SPTM/SGM remains the same except the agent used to generate the training data is varied between FS, RS, and EC.
For the EC variant, the training data is composed of trajectories that were recorded during the training of the exploration module.

\bgroup
    \def\arraystretch{1.1}
    \setlength{\tabcolsep}{4pt}
    \begin{table}[t]
        \footnotesize
        \begin{center}
            \begin{tabular}{ c c c | c c c c } 
             \toprule
            \multicolumn{3}{c|}{Method Used} & \multicolumn{4}{c}{\textit{Test-1}} \\
            Populate Graph & Train Networks & Evaluation & Easy & Medium & Hard & Hardest \\ 
             \midrule %
             - & FS & Controller & 43\% & 12\% & 0\% & 0\%  \\ 
             - & RS & Controller & 29\% & 2\% & 1\% & 0\% \\
             - & EC & Controller & 39\% & 6\% & 0\% & 1\% \\
             \midrule %
             FS & FS & SPTM & 32\% & 4\% & 0\% & 0\%  \\ 
             RS & RS & SPTM & 32\% & 6\% & 0\% & 0\%  \\ 
             \underline{EC} & \underline{FS} & \underline{SPTM} & \textbf{48}\% & \textbf{16}\% & 2\% & 0\%  \\ 
             EC & EC & SPTM & 46\% & 12\% & \textbf{4}\% & \textbf{4}\% \\
             \midrule %
             FS & FS & SGM & 41\% & 8\% & 0\% & 0\% \\
             RS & RS & SGM & 24\% & 1\% & 0\% & 0\% \\
             \underline{EC} & \underline{FS} & \underline{SGM} & 43\% & 3\% & 0\% & 0\% \\
             EC & EC & SGM & 29\% & 0\% & 0\% & 0\% \\
             \midrule %
             \multicolumn{3}{c|}{MSS} & 23\% & 9\% & 1\% & 1\% \\
             \multicolumn{3}{c|}{\slm [ours]} & 85\% & 59\% & 62\% & 50\% \\
             \bottomrule %
            \end{tabular}
        \caption{(\textit{Fixed spawn}) The success rates of ablations of SPTM and SGM baselines on the \textit{Test-1} \vizdoom{} map. For each difficulty level, we bold the success rate of the best-performing ablation baseline for emphasis. We also include the success rates of MSS and our method \slm for reference.}
        \label{table:ec-ablations}
        \end{center}
        \vspace*{-0.25in}
    \end{table}
\egroup

We then evaluate the underlying baselines initially described in \Cref{sec:baselines}, visual controller\footnote{The visual controller baseline does not use a high-level graph for planning.}, SPTM, and SGM, with ablations of the methods used to populate the graph and train the networks.
We report their success rates averaged over 5 random seeds in \Cref{table:ec-ablations}.
The EC (populate graph) + FS (train networks) + SPTM/SGM baselines (underlined) are the ones reported in the main paper in \Cref{table:fixed-vizdoom}.
From these experiments, we make the following observations.
First, using EC-generated trajectories for populating the graph can improve SPTM's performance on longer-horizon goals.
This is expected as the exploration bonus from EC supports greater coverage of more distant areas of the state-space, which thereby enables graph planning to distant goals.
Second, we find that the FS-generated trajectories, while limited in their coverage of distant areas, can outperform RS and EC on the Easy and Medium difficulties.
We hypothesize that this is due to how \textit{fixed spawn} start-goal pairs in evaluation share the same starting location as in training.
With the FS trajectories, the training of the networks is skewed towards goals closer to the starting location.
Conversely, RS trajectories suffer by having episodes start from a different location than the start location of the evaluation setting.

\begin{figure}[!t]
    \centering
    \hfill
     \begin{subfigure}[b]{0.31\linewidth}
        \centering
        \includegraphics[width=\linewidth]{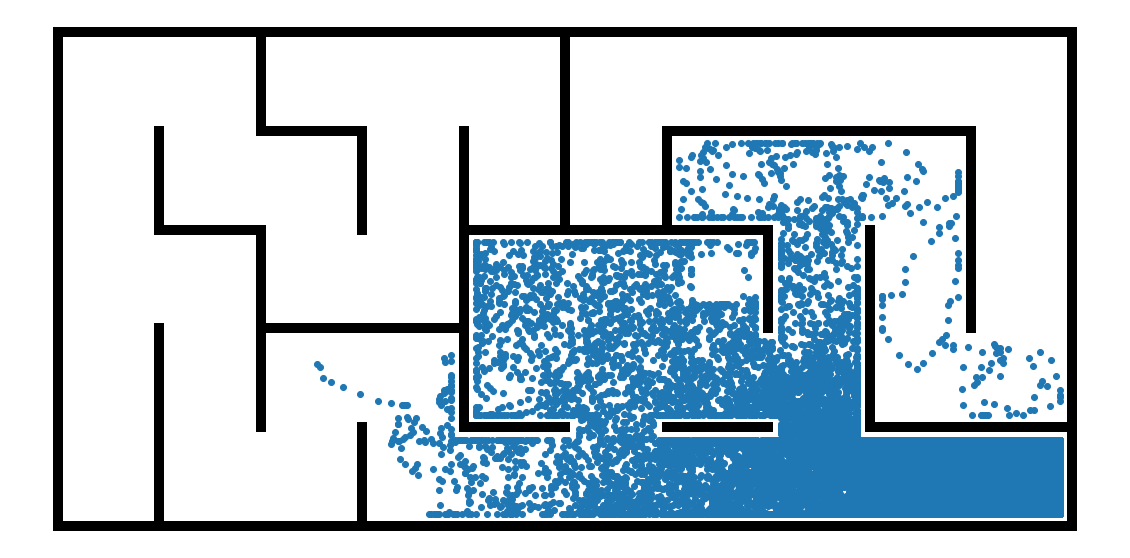}
        \caption{FS}
        \label{fig:sgm-trajs-fixed}
    \end{subfigure}
    \hfill
     \begin{subfigure}[b]{0.31\linewidth}
        \centering
        \includegraphics[width=\linewidth]{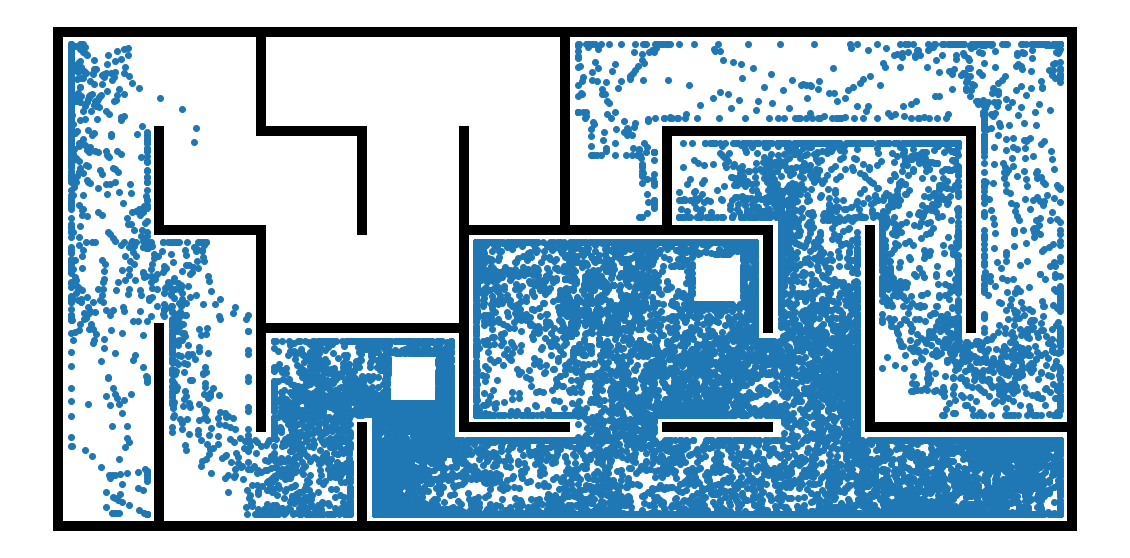}
        \caption{EC}
        \label{fig:sgm-trajs-ec}
    \end{subfigure}
    \hfill
     \begin{subfigure}[b]{0.31\linewidth}
        \centering
        \includegraphics[width=\linewidth]{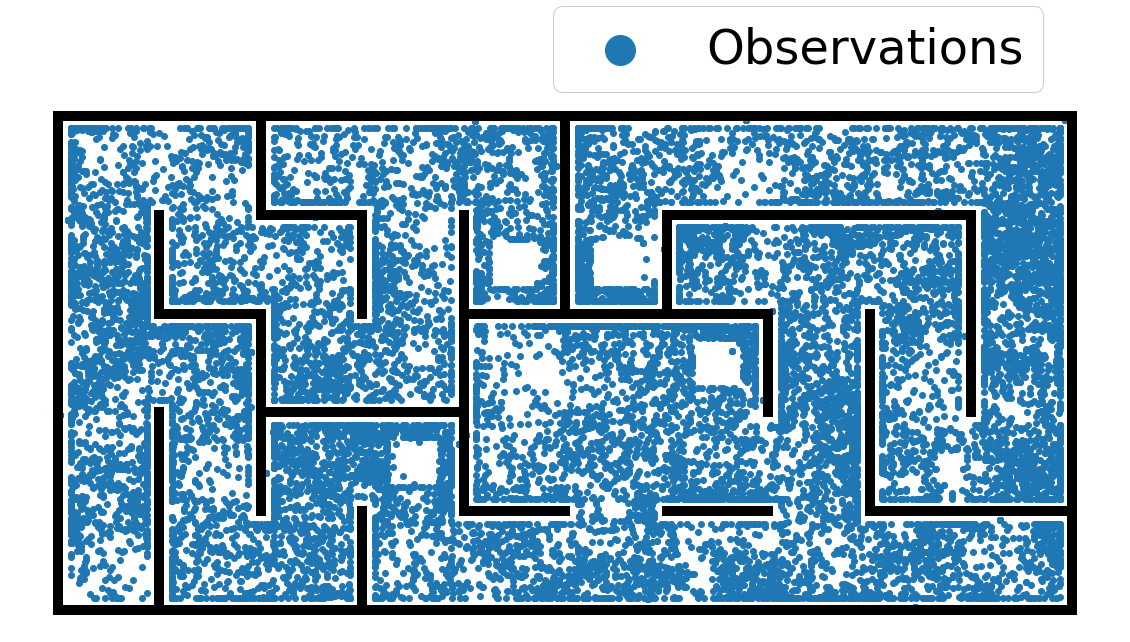}
        \caption{RS}
        \label{fig:sgm-trajs-random}
    \end{subfigure}
    \hfill
    \caption{State coverage of the replay buffers built by the FS, EC, and RS agents.}
    \vspace{-10pt}
    \label{fig:sgm-trajs}
\end{figure}

We also visualize the state coverage of the replay buffers used for populating the graph for the FS, EC, and RS agents in \Cref{fig:sgm-trajs}.\footnote{The four distinct white squares are caused by the presence of peripheral in-game objects.}
The EC agent is able to store observations from more distant locations in comparison to the FS agent.
As expected, the RS agent provides comprehensive coverage of the entire map.

\clearpage
\section{Additional Details: Traverse Algorithm}
\label{sec:appendix-algorithm}

Here is the Traverse algorithm used for localizing the agent to the nearest landmark ($\landcurr$) based on SFS and traversing to the next target landmark ($\landtarget$) on the landmark path with the goal-conditioned policy($\pi_l$).
The procedure is repeated until either the target landmark is reached or the ``done'' signal is given indicating that the episode is over.
This algorithm is referenced in \Cref{alg:overall2}.
\vspace{12pt}

\begin{algorithm}[H]
	\caption{\textsf{Traverse}}
	\label{alg:traverse}
	\begin{algorithmic}[1]
		\INPUT{$\pi_l, \landtarget$}
		\OUTPUT{$\tau_\text{traverse}, \landcurr$}
		\STATE $\tau_\text{front}\gets\emptyset$
		\WHILE{$\landcurr != \landtarget$ or env not done}
			\STATE $(s, a, r)\sim \pi_l(\cdot; \landtarget)$
			\STATE $\landcurr\gets \argmax_{l\in L}\sfs_{\theta}(s, l)$ \hfill\COMMENT{localize agent}
			\STATE $\tau_\text{traverse} = \tau_\text{traverse} \cup (s, a, r)$
		\ENDWHILE
		\STATE \textbf{return} $\tau_\text{traverse}, \landcurr$
	\end{algorithmic}
	\smallskip
\end{algorithm}

\section{Societal Impact}
\label{sec:appendix-societal-impact}

Our Successor Feature Landmarks (\slm) framework is designed to simultaneously support exploration, goal-reaching, and long-horizon planning.
We expect for our method to be applicable to real-world scenarios where an autonomous agent is operating in a large environment and must complete a variety of complex tasks.
Common examples include warehouse robots and delivery drones.
\slm can improve the efficiency and reliability of these autonomous agents, which offer potential benefit to human society.
However, these autonomous systems may also be built for more malicious purposes such as for constructing weaponry or conducting unlawful surveillance.
In general, these potential harms should be carefully considered as we begin to develop autonomous agents and pass legislation that will govern these systems.

\end{document}